# Towards dynamic narrow path walking on NU's Husky

A Thesis Presented

by

**Kaushik Venkatesh Krishnamurthy**

to

**The Department of Mechanical and Industrial engineering**

in partial fulfillment of the requirements

for the degree of

**Master of Science**

in

**Mechanical Engineering - Concentration in Mechatronics**

**Northeastern University**
**Boston, Massachusetts**

December 2023

*To my family and friends and everyone along the way.*

# Contents

# List of Figures

# List of Tables

# List of Acronyms

**DoF** Degrees of Freedom.

**HROM** Husky Reduced Order Model.

**LTI** Linear Time Invariant.

**LQR** Linear Quadratic Regulator.

**MPC** Model Predictive Control.

**GRF** Ground Reaction Forces.

**HZD** Hybrid Zero Dynamics.

**ZMP** Zero Moment Point.

**CoM** Center of Mass.

**CoP** Center of Pressure.

**HF** Hip Frontal.

**HS** Hip Sagittal.

# Acknowledgments

I want to express my gratitude to my advisor Dr. Alireza Ramezani for his continued support and guidance throughout this thesis. I also want to express gratitude to Dr. Eric Sihite for his assistance. I also thank my fellow lab mates Shreyansh, Adarsh, Chenghao, Bibek, Kruthika, Dharsan and others for their invaluable help in conducting physical testing, dealing with hardware components and mechanical design, and for their stimulating discussion on the research. I would also like to thank my friends Manickam, Subha, Poornima, Kaushik S, Rakshitha, Kishan, Abhinav, Naveen, Pathy, Aish, Pathy , Selva, Evan, Smruti, Bhushan, Jessey, Miles and everyone else who supported me in this endeavour. Finally, I would like to express my gratitude to my family, who believed in me and are the biggest encouragement behind all my hard work.

# Abstract of the Thesis

Towards dynamic narrow path walking on NU's Husky

by

Kaushik Venkatesh Krishnamurthy

Master of Science in Mechanical Engineering - Concentration in Mechatronics

Northeastern University, December 2023

Dr. Alireza Ramezani, Advisor

Dr. Rifat Sipahi, Thesis Reader

This research focuses on enabling Northeastern University's Husky, a multi-modal quadrupedal robot, to navigate narrow paths akin to various animals in nature. The Husky is equipped with thrusters to stabilize its body during dynamic maneuvers, addressing challenges inherent in aerial-legged systems. The approach involves modeling the robot as HROM (Husky Reduced Model) and creating an optimal control framework using linearized dynamics for narrow path walking. The thesis introduces a gait scheduling method to generate an open-loop walking gait and validates these gaits through a high-fidelity Simscape simulation. Experimental results of the open-loop walking are presented, accompanied by potential directions for advancing this robotic system.

# Chapter 1

# Introduction

      Biologically inspired robots have been the subject of major interest in the research of mobile robotics especially in the past few decades. Biological systems, having evolved to address specific challenges in nature, offer valuable insights for engineering solutions through bioinspiration. Mobile robots, especially those incorporating bioinspired elements, grapple with challenges inherent in their design, including restricted energy density, mass limitations, and under-actuation. Animals have adeptly navigated such challenges by leveraging lightweight musculo-skeletal structures, providing the requisite energy density for effective actuation. These structures are a result of evolutionary design and we wish to harness this evolutionary design in our robots. Some of the most celebrated works in bio-inspiration can be found within the research team of *SiliconSynapse* Labs, with a revolutionary lightweight bat inspired robot *Aerobat* [39] [46] [27] [54]. The robot uses a skeletal structure driven by a single motor to aid flapping motion that closely mimics that of a bat [49] [15]. Another robotics system from the same team that is bio-inspired is COBRA. COBRA is a morphing robot to mimic a snake's motion. But the robot also has an ability to reshape it's links to become a tumbling robot to explore craters on the moon [43].

      The inception of modern legged systems can be traced back to the early 1980s when Raibert [36] constructed a hopping robot, followed by a walking robot with hydraulically actuated legs. By the turn of the century, bipedal systems like the Honda Humanoid Robot [19] and ASIMO [42] were introduced. A different approach emerged, shifting away from the cumbersome design of fully actuated-legged robots like ASIMO. Instead, focus shifted towards underactuated and passive walkers [14], showcasing the efficiency of legged systems, particularly in terms of low transport cost. But regardless of actuation, the ability to balance during both simple and complex dynamic maneuvers is an inherent trait in humans and other bipedal/quadrupedal animals, facilitated by

powerful proprioceptive and exteroceptive sensors. However, replicating this capability in robots presents challenges. Estimating position, orientation, and instantaneous velocities becomes intricate proprioceptively, and the computational expense increases significantly when relying on exteroceptive sensors. Difficulties also emerge as a result of very ability of legged systems to only manipulate through the environment with the help of the robot's contact with the ground, which in fast dynamic maneuvers can be an extremely small duration [9]. Examples of studies where appendages like tails [11] [29] [50], limbs [2], reaction wheels [26] and even wings [38], [40] [20], [55], to generate opposite momentum, exhibit dynamic maneuvers, but these come at the cost of increased hardware complexity and weight on a robot the size of Husky.

A recent theme of bio-inspired robotics is the ability to be multi-modal. The remarkable multi-modal capabilities of certain animals to navigate various terrains is also captivating a lot of attention. Some of these include Chukar birds which have the ability to walk up high inclines [12] and perform agile maneuvers such as rapid walking, leaping and jumping using both their legs and wings. In situations where only one mode of locomotion is not sufficient, multi-modal locomotion can be a solution. Taking legged and aerial for example, which have have their own advantages and disadvantages. Aerial systems have the ability to traverse large distances very fast albeit with a very high cost of transport. On the other hand, legged systems can walk with a very low cost of transport but at the expense of time and inability to cross difficult terrain. When designing aerial-legged robots and designing controllers, one must face antagonistic requirements, i.e to be able to fly the robot must be lightweight and to be able to walk the robot must be able to have strong actuators that can take it's weight. Attempts to circumvent this from previous research in *SiliconSynapse* Labs have been explored by [32] [48] and [6]. Optimization based methods can become very expensive to satisfy some of the trajectory constraints in legged-aerial robots, and Dangol et al. address these in [9].

Some of the celebrated works in multi-modal robotics began as early as in 1980's [21] to the current day where there are multi-modal wheeled aerial, legged-aerial, and amphibian robots [58]. M4 [53], a multi-modal wheel-legged-aerial robot is a testament to the strengths of multi-modal locomotion, where the robot is able to repurpose it's appendages for different types of locomotion [30] .Path-planning then becomes a separate avenue of research, with some successful attempts to perform this on M4 in simulation and real have been exhibited in [52][51]. Path planners need to be able to distinguish the ability of the robot to move over a specific obstacle with the right choice of locomotion modality. Some of these works are shown in Fig. 1.1.

LEONARDO [25] and Harpy [41] are some examples of legged aerial robots which make use of their legs as well as thrusters to stabilize their gaits and overcome obstacles such as a flight

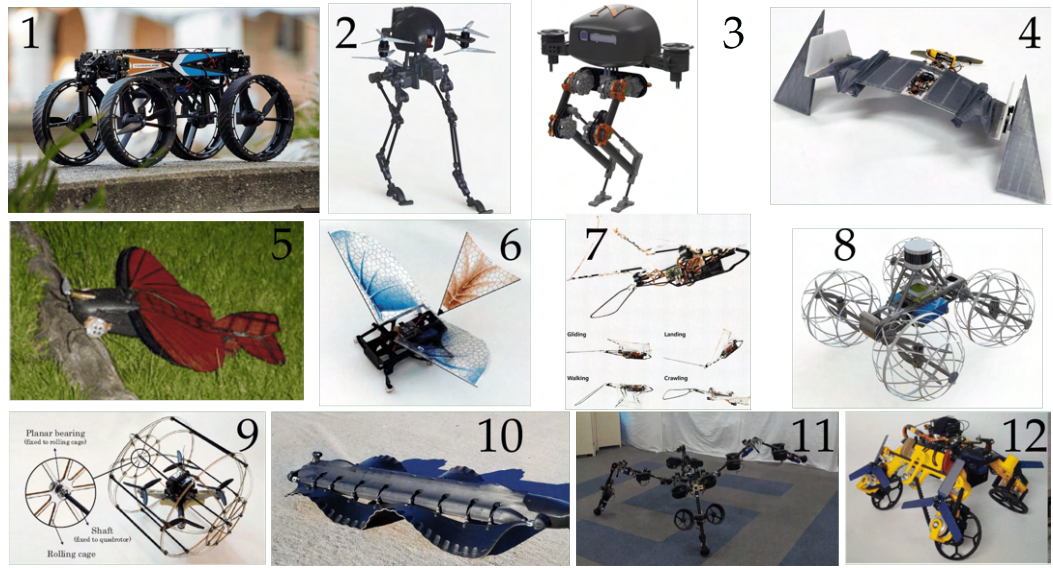

Figure 1.1: A collection of celebrated multimodal robots - 1) M4 [53], 2) LEONARDO [25], 3) Harpy [34] 4) DALER [3] 5) MMALV [22], 6) DASH + Wings [33] 7) Pteromyini [44] 8) Drivocopter [24] 9) HyTAQ [23] 10) Velox [58] 11) SPIDAR [62]12) FCSTAR [10], figures borrowed from respective references

of stairs. While LEONARDO used quasi-static gaits and static gaits, Harpy can go further with the addition of custom BLDC actuators and the ability to perform dynamic gaits and jumps [4]. Some other successful multi-modal robots that include Morphing Micro Air-Land Vehicle (MMALV) [22], which was one of the first studies to use wings and ground mobility. This was followed by DASH+wings [33] a bio-inspired cockroach robot that was able to use it's wings to run up inclines, much akin to the aforementioned Chukar birds. *Pteromyini* a bio-inspired legged and aerial robot that uses legs to jump a soft membrane as wings to help glide mimicking the flying squirrel of the same name. Adding to the list of bio-inspired multi-modal robots is SPIDAR [62], a quadruped robot that is fitted with vectorable thrusters on its legs which gives it the ability to seamlessly transition between walking and flying. DALER [3], a robot that is inspired by the properties of a bat *D. rotundus*, which uses its limbs to both walk and fly. The robot utilises a skeletal structure that can morph it's foldable wings to use it as limbs to walk.There are examples of rotary wing multicopters which have multi-modal capabilities such as the Drivocopter [24] and HyTAQ [23] and the wall climbing FCSTAR [10] use a form of wheeled aerial locomotion.

From all the above research, what can be concluded is the ability to fly is especially desired

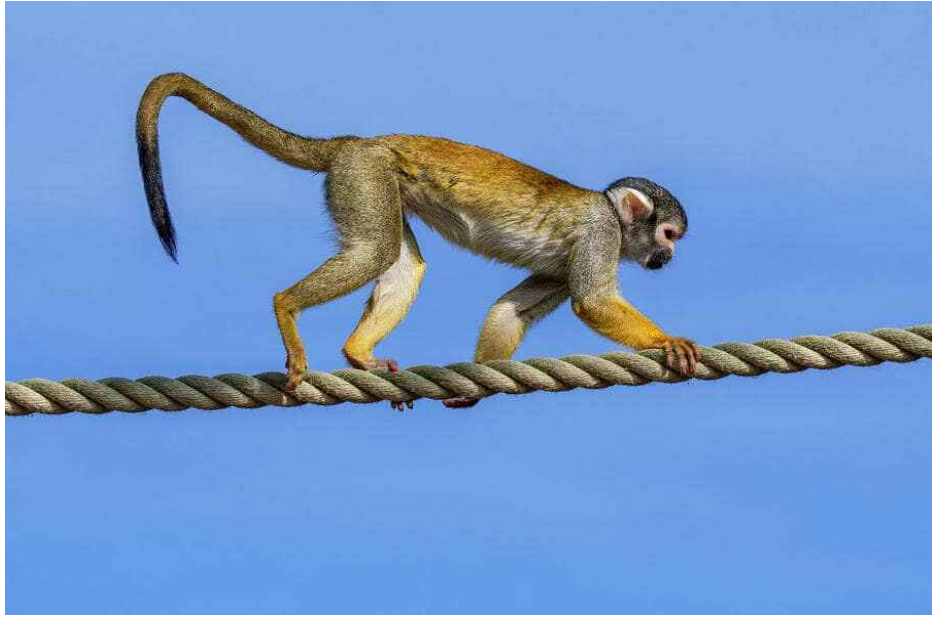

Figure 1.2: Squirrel Monkey walking on slackline cable. This specific species of monkeys use their tails extensively to aid arboreal quadrupedalism [61] [60]

because flying is not encumbered by most obstacles. This combined with the ability to posses dynamic terrestrial locomotion is a quality that can be very desirable for mobile robots.

## 1.1 Narrow Path Walking : History and Motivation

Narrow path walking or balancing on a beam, involves a high level of dexterity and balancing and is something that certain animals like cats [56] do with ease. Cats have the ability to twist their spine and use their tail to generate corrective forces and momentum to balance. An example of this can be seen on Fig. 1.4. The squirrel monkey Fig. 1.2 also uses it's tail extensively to balance itself on something like a very thin cable. Legged-robots however have this much more difficult as the only point where the momentum of the body can be manipulated is at the and point of contact. To overcome this, some robots implement methods to external mechanisms to generate momentum. One research shows a quadrupedal robot balancing while crossing a balance beam by generating momentum through reaction wheel actuators [26]. LEONARDO [25] exhibits the ability to slackline by manipulating forces from the thrusters on the robot to balance the body. Another way to do this, is by mapping the kinematics and dynamics of the robot using reduced order models such as a double inverted pendulum and by applying methods to balance it. Gonzalez et al. [16] and [35] exhibit line walking in simulation and in real-life. The study specifies the importance of accurate torque measurements, the problem of sensitivity to disturbances and how unmodelled compliance can introduce underactuated DOF's. IHMC's Atlas robot [18] [17] also exhibits impressive line balancing capabilities on extremely thin beams using a whole body control method. There has also been attempts at using Reinforcement based methods have also been investigated by Xie et al. [59] where a RL policy is trained to make a quadrupedal robot to perform a variety of maneuvers including balance beam locomotion.

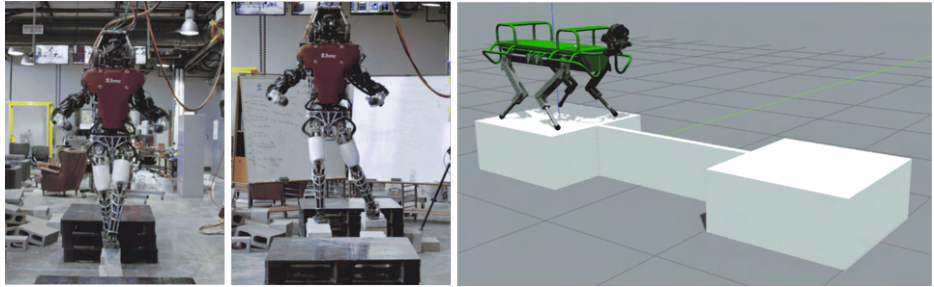

Figure 1.3: IHMC's Atlas robot exhibiting narrow path walking with crossover steps [18] and simulation of legged performing narrow path walking wiht point feet [16]

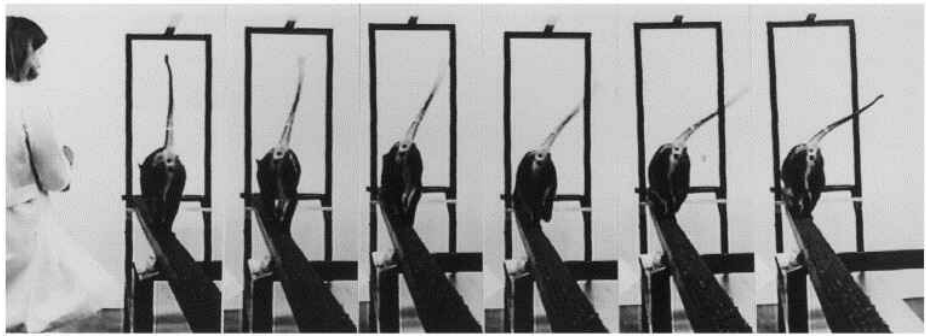

Figure 1.4: Cat exhibiting narrow path walking on a balance beam and simultaneously using its tail to balance the body [56]. The tail helps in generating an external moment/torque.

    Line walking, in general as an ability for legged robots, can improve the robustness of robots while walking over difficult scenarios. An application that is particularly attractive is the conditional monitoring of power transmission lines, which are exposed to the elements and can deteriorate over time. Downed wires can cause extended downtimes and can cause a lot of strain to the power-grid. Another application where robots have been considered with specific attention is search-and-rescue missions during disasters, where robots have to deal with rubble, and uneven

terrain. The ability for robots to balance on thin beams can become highly transferable and suitable for such missions.

## 1.2 Objectives and Outline of Thesis

The main motivation for my thesis project is to fill the research gap in the field of multi-modal legged-aerial robot locomotion with the help of Northeastern University's Husky platform. The thesis also focuses on manifesting a specific type of walking trajectory which allows the robot to walk on very narrow platform with sufficient frontal plane stabilisation. This calls for a design and assembly of integrate thrusters to the robot which helps stabilise the robot. To validate the ability of these walking trajectories a simulation model is employed on Simulink®Simscape before actual implementation on the robot. The high level objective of this design to help make progress towards aerial-legged locomotion and be able to harness the most out of it's advantages. Quadrupedal locomotion that employs quasi-static gaits (like three point contact walking) is a simpler problem, because with proper weight distribution can have the robot can have its the projection of the COM falls within the support polygon. In dynamic gaits such as the two point contact walking where the support polygon at most times is very small, multi-modal locomotion can provide stability during these walking. Multi-modal locomotion can open avenues for multiple different quadrupedal maneuvers and my thesis investigates the ability of one such maneuver. The specific goal is then to evaluate the performance of a stable walking trajectory and an inexpensive control schematic to meet this.

Initially the Husky Carbon platform is introduced in Chapter 2 by describing the design, components and control challenges of the system. Further, an overview of related design concepts and control design paradigms will be discussed. Chapter 3 covers the reduced order dynamic modeling of the system using Euler-Lagrangian modelling. Chapter 4 presents an optimal control strategy using locally linearized dynamics and some initial results on MATLAB®. This chapter also covers the trajectory generation startegy using Bézier polynomials. Chapter 5 presents the Simscape simulation setup to validate the gait generation methods and results. In Chapter 6, the physical testing procedure along with the obtained results are discussed.

## 1.3 Contributions

My thesis delves into the current progress of the project, outlining both my contributions and a prospective strategy to advance the platform. I built upon the groundwork laid by my peers, who had already established the structure of the Husky MATLAB® firmware, along with the integration of a host computer and power electronics. Specifically, my contributions to the firmware involved updating the Simulink® real-time code to incorporate a scheduled gait generation method using Bézier control points. The hardware for the Husky had been developed before my program commenced, and my distinct role included wiring the power electronics and integrating the propulsion unit with EDFs and other sensors. In the simulation phase, I constructed a high fidelity simulation model of the Husky on Simulink® Simscape Multibody, showcasing thruster-assisted legged locomotion and validation of the open loop gaits generated. I also present a control strategy for Husky using an MPC using local linearization of the reduced order dynamics modelling of Husky. Beyond simulation, I, alongside the invaluable assistance of my colleagues, actively leaded the experimentation and data collection on the physical Husky platform.

# Chapter 2

# The Husky Carbon Platform

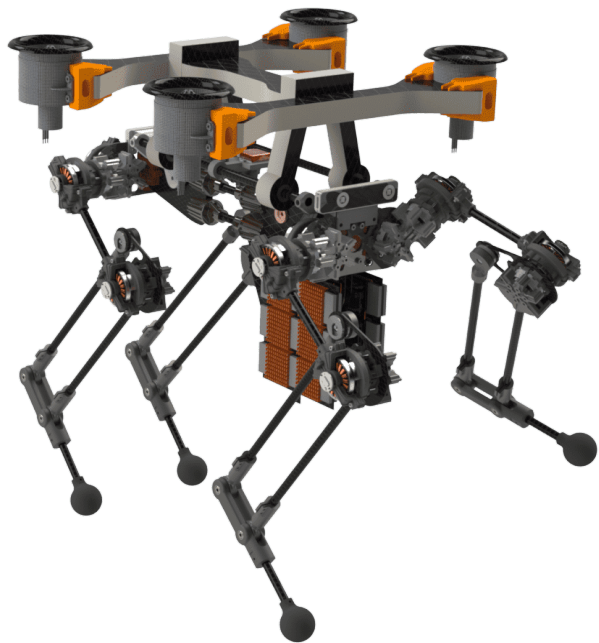

Figure 2.1: Solidworks render of Husky showing the major components

## 2.1 Brief Overview Of Husky Carbon and its components

Husky Carbon, is a quadrupedal robot with multi-modal capabilities, custom designed and fabricated at Northeastern University's *SiliconSynapse Labs*. Husky Carbon, along with Husky Beta [57], Harpy and M4 join the suite of robots with multimodal capabilities. Husky Carbon, hereby referred to just as Husky, stands at 1.5 ft wide and 3 ft tall. Most of the skeletal structure of Husky was built using carbon fiber plates and continuous carbon fiber reinforced additive manufactured components printed using Markforged 3D-printers and their proprietary **Onyx** thermoplastic. The design of Husky was derived from extensive research in generative design to reduce mass-moment and inertia and TCOT (Total Cost of Transport) [41] which resulted in a front-heavy robot. Two pairs of legs are attached to two pairs of pelvis plates, which house the hip joints. The leg offers three degrees of freedom across two distinct planes: the frontal and the sagittal plane. The hip-frontal (HF) joint facilitates movement in the frontal plane, while the hip-sagittal (HS) works in tandem with the knee (K) joint to maneuver the leg in the hip sagittal plane. All the 12 joints are actuated by T-motor Antigravity 4006 brushless motors, with the motor output transmitted through a Harmonic drive. Harmonic drives enable precise joint movement with minimal backlash and allow for back-drivability. The motor and gearbox housings were intentionally designed to be exceptionally lightweight, with the components embedded during the printing process. This approach not only integrates the components seamlessly into the design but also significantly reduces the need for numerous fasteners.

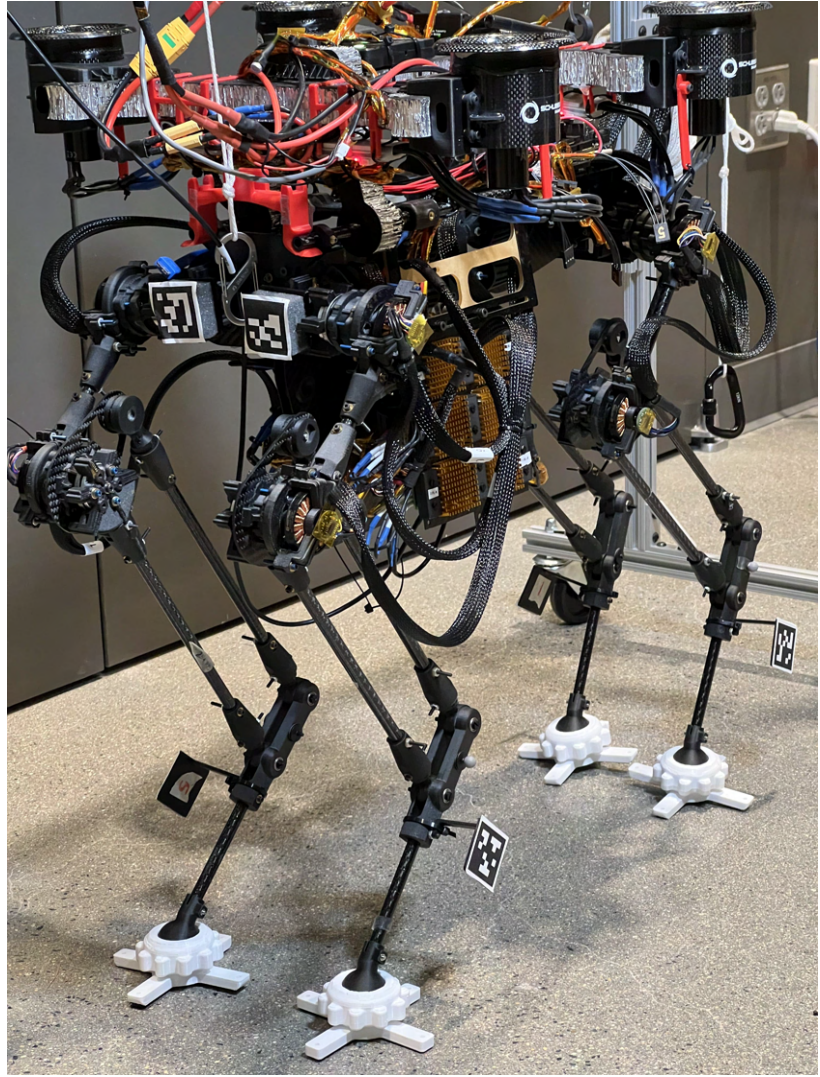

Figure 2.2: Actual Husky Carbon platform. The robot stands 1.5 ft wide and 3 ft tall

To power the 12 actuators, the robot uses 12 ELMO gold twitter solo amplifiers, which are advanced high power-density servo drives that are built into a really small form factor. The amplifiers are wired to using a daisy chain method where all 12 amplifiers are connected in a series fashion and the first amplifier on the chain is connected to the Speedgoat master realtime machine, also equipped with IO581 for Serial communication, using EtherCAT. EtherCAT, or **Ether**net for **C**ontrol and **AuT**mation systems, provides a protocol which facilitates merging the capabilities of Ethernet and real-time operating systems. The realtime machine sends the amplifiers control commands also get position feedback from the drives using incremental magnetic encoders. The amplifiers are isolated away from the actuators and are separated into 2 racks of 6 amplifiers each and then mounted onto either side of the robot. The host PC running MATLAB® is connected to the Speedgoat Realtime machine using EtherCAT.

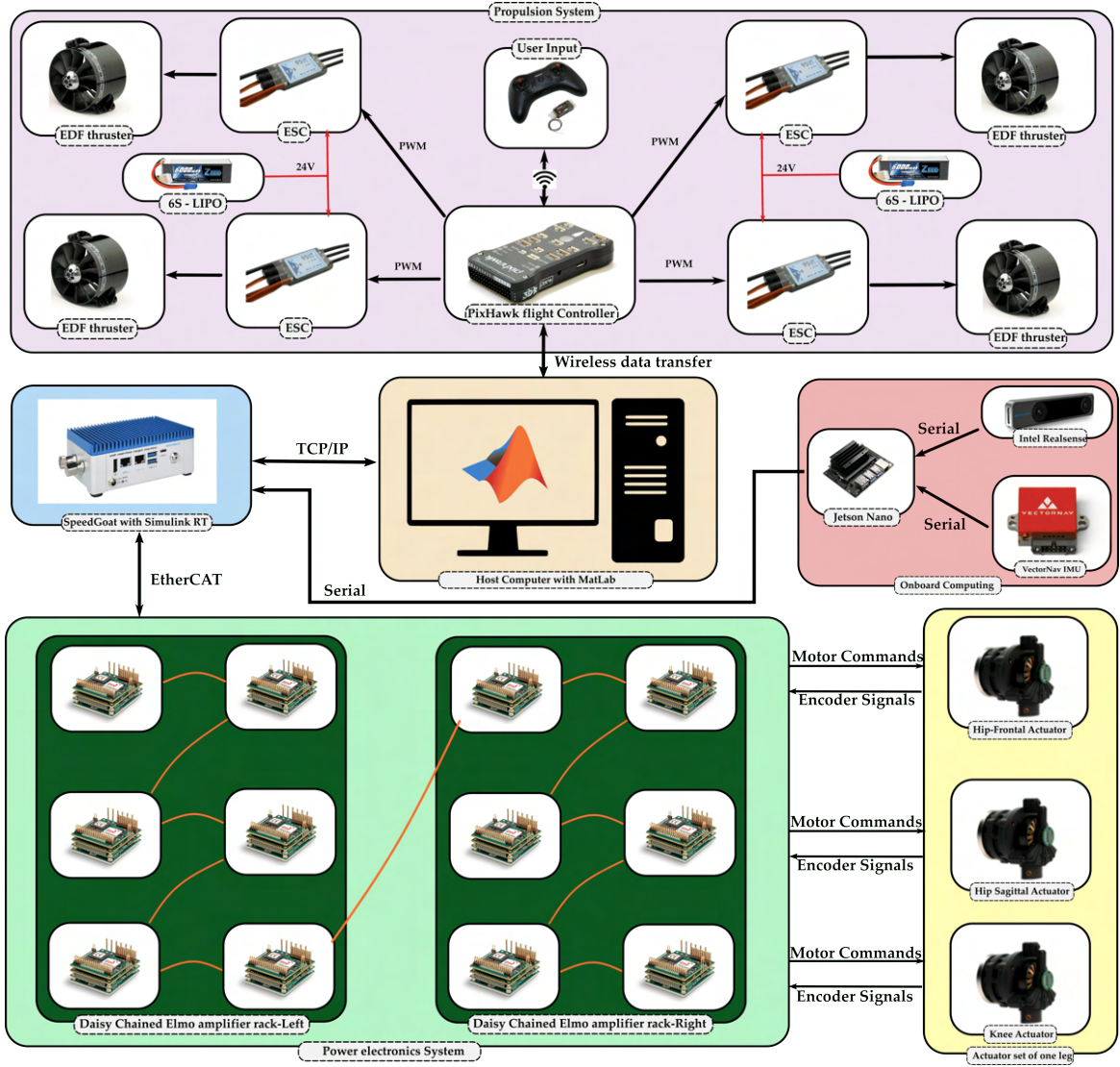

Figure 2.3: Illustrates an overview of Husky's communication system

The propulsion system is fitted with 4 Schubeler Electric Ducted fans (EDF) that provide approximately 8 kgf of thrust and is built with lightweight hexagonal aluminum composite structure sandwiched between carbon fiber plates. The particular selection and effectiveness of this sandwich structure is discussed in the design of Harpy [34], another multi-modal legged robot from *SiliconSynapse Labs*. The thrusters are mounted on fiber-reinforced 3D-printed mounts and an embedding process is used to attach them on to the composite structure. Here the composite is made out of 3 parts, a central part that holds the four thrusters and two identical parts that is attached to the central structure with the help of epoxy. The propulsion unit is then assembled on the robot with the help of two carbon fiber rods that run along the length of the robot. The inherent design of the propulsion unit again allows for minimal usage of fasteners and as result reduces the weight of Husky.

The EDF's are powered by 2 6S-LiPO batteries and are driven by 4 YGE 95 LVT ESC's. The control commands for the EDF's are sent to the ESC's using a Pixhawk flight controller, which obtains user inputs from a Radiolink controller and receiver on the Pixhawk. A wireless transfer between to Pixhawk and the host PC is used to collect thrust, IMU and current readings. The figure Fig. 2.3 gives an overview of the communications between the different systems on Husky.

For the purpose of high-level navigation and planning Husky is equipped with a Jetson Nano, an Intel realsense t265 tracking stereocamera and a Vectornav VN200 IMU. Autonomous path planning and exploration for multimodal robots adds an additional layer of complexity as now decisions about when to move transform between different locomotion modalities needs to be made based on the robot's surroundings. Attempts to solve this problem have been made in by Sihite et al. [52] on Husky simulations and Rajput [37] on M4 simulation with successful demonstrations of the planning for multimodal locomotion.

| Component | Name |
| --- | --- |
| IMU | VN-200 |
| Stereo Camera | Intel Realsense |
| Actuators | T-motor Antigravity 4006 brushless |
| Harmonic Drives | CSF-11-30-2A-R |
| Servo Drives | ELMO gold twitter |
| Encoders | RLS RMB20 |

Table 2.1: Components on Husky

## 2.2 Challenges in multi-modal locomotion on Husky

The Husky robot faces new and tough control challenges that are previously not seen on most robots. To accommodate aerial locomotion the weight of the robot is designed to be as minimal as possible. This is mainly because of the antagonistic nature of aerial-legged locomotion. The robot must be lightweight and the thrusters must be able to provide a sufficient thrust-to-weight ratio to perform any aerial maneuvers. On the other hand, while walking, the legs have the largest effect in the transfer of momentum to the robot's body through the contact points. To also perform dynamic aerial-legged maneuvers, a robust control method to have a stable posture and movement is required. This is especially hard because, in the current iteration of Husky, in the absence of thrust vectoring, body posturing comes into play to manipulate the external wrench force on the robot. This is an active field of research and also proves to be a promising avenue to explore in the field of aerial-legged robotics.

To reduce the total mass moment of inertia laterally, the legs of Husky are placed very close to each other. As a result, the support polygon of Husky is small and almost non existent in the presence of point feet making the stability margins very small. The inherent stability then found in quadrupeds is lowered in Husky further increasing the challenges for gait design and control. Another added difficulty is foot trajectory generation to avoid self collisions. Gaits also need to be properly timed to make the sure the legs impact the ground at the right time for different types of gaits.

Reducing the mass and stiffness of the robot not only decreases its overall weight but also diminishes its rigidity. Consequently, Husky also becomes more prone to compliance issues, especially when specific components are subjected to high loads. Addressing this challenge poses both mechanical and control engineering challenge as we try to circumvent compliance by trying to model it as an uncertainty in the system, but this specific effort is out of the scope of my thesis.

## 2.3 Narrow path walking in Husky

Narrow path walking is possible in Husky mainly due to the fact that the legs have significant space to swing in the frontal plane and this is facilitated by the hip frontal joint. This allows the leg to cross over each other and this creates the ability to walk in a line.

For assessing stability while performing narrow path walking, we make sure that the robot's leg maintains friction cone constraints. This is even more so important in narrow path walking as the leg contacts the ground in an angle which may increase the horizontal component of the contact force. The friction cone constraints can be formulated as follows:

$$\begin{aligned}
\frac{||\boldsymbol{F}_h||_2}{||\boldsymbol{F}_z||_2} &< \mu \\
F_z &> 0 \\
||\boldsymbol{F}_h||_2 &= \sqrt{F_x^2 + F_y^2}
\end{aligned} \quad (2.1)$$

where here, the terms $\boldsymbol{F}_h$ and $\boldsymbol{F}_z$ denote the horizontal forces and the unilateral normal force respectively. $\boldsymbol{F}_h$ can be composed of $\boldsymbol{F}_x$ and $\boldsymbol{F}_y$ which are the components of the horizontal force on the basis axis of the ground surface and here $||.||_2$ is the Euclidean norm. To ensure stability at the point of contact, we add these constraints in the while modelling and simulating our robot. Often the friction cone is linearized as polytope structures like pyramids.

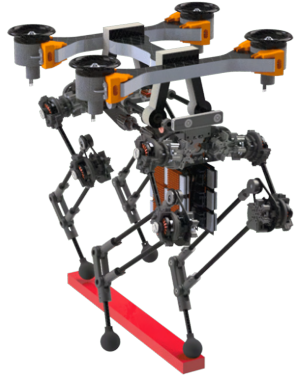

Figure 2.4: Illustrates a render of Husky walking on a narrow path with a two point contact gait

To stabilize the body pose, an external thruster force, modelled as a wrench in the modelling and simulations, is manipulated to compensate for any roll, pitch and yawing errors.

### 2.3.1 Gaits

We plan to employ a two-point contact dynamic gait on a narrow path. This gait more technically known as a diagonal sequence diagonal-couplet (DSDC) because the diagonal limbs are related in time as a pair. This strategy means that at any instance the robot has only two-points of contact on the ground. At any point during the gait two diagonally opposite pairs of legs will perform the same action, i.e swinging or stancing. The timing of a two-point contact walking is such that the two diagonally opposite pairs of legs are always fully out of phase with each other. And only for a very short period of time (infinitesimally small in an ideal scenario), when one pair is just about to start the swing phase and the other pair is about to finish the swing phase, all the 4 feet will be on the ground.

This type of gait, commonly known as trotting and is exhibited by dogs, cats, goats, horses and some other quadrupeds as an intermediate between a slow crawling gait and an all out gallop. We employ the trotting gait in Husky by generating trajectories for the leg using Bezier control points. The thrusters on the robot provide forces to compensate for the roll, pitch and yaw errors. For some other animals such as chameleons have been found to adopt a Lateral Sequence Lateral-Couplet (LSLC) gait, where legs on either side of the body are coupled in time.

# Chapter 3

# Husky Reduced Order Modelling

Modelling the entire dynamics of quadrupedal robots using a high fidelity model and by assessing every single degree of freedom can become very tedious, and costly to compute numerically in simulations. As a result, there are advantages to modelling Husky with reduced order dynamics and here I present the Husky Reduced Order Model or HROM. The HROM has been used in a few studies with successful simulations and controllers [7] [5] [28]. Husky's HROM has 2 levels of complexity where the HROM-Level 1 has a single rigid body with a leg structure that is similar to the leg structure of the full model while the mass of the legs is concentrated as point masses on the hip and knee joints. Whereas on HROM level 2, the dynamics is further simplified with massless variable length linkages from the hip and the hip being concentrated as a spherical joint with 2 primitive rotations. An illustration of the HROMs are shown in Fig. 3.1

For the purposes of this thesis we consider the dynamics of the HROM-Level 2, hereby referred to as simply HROM. The equations of motion of the HROM can be derived using Euler-Lagrange dynamics formulation. The position of the legs can be defined as function of the spherical joint primitives, namely $\phi, \gamma$ and the length of the leg $l$. The position of the body can be defined with a position $\boldsymbol{p}_B \in \mathbb{R}^3$ and a rotation in SO(3) with $\boldsymbol{R_B}$. The generalized co-ordinates of the body and leg can then be defined as follows.

$$\begin{aligned} \boldsymbol{q_L} &= [\phi_i, \gamma_i, l_i] \\ \boldsymbol{q_B} &= [\boldsymbol{p}_B^\top, \boldsymbol{r}_B^\top] \\ i &\in \mathcal{F}, \end{aligned} \quad (3.1)$$

where $\mathcal{F} = [FR, BR, FL, BL]$ and $\boldsymbol{r}_B \in \mathbb{R}^3$ contains the elements of the rotation matrix $R_B \in$

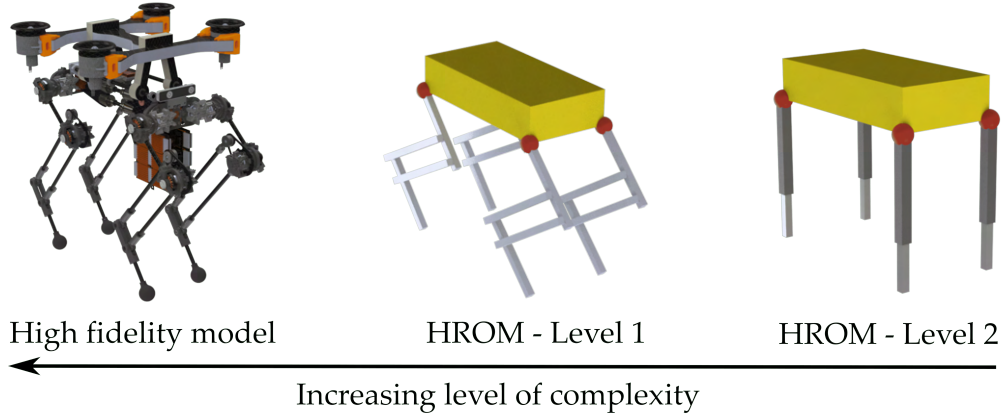

Figure 3.1: HROM- Increasing levels of complexity that can be added by adding more information about the robot while modelling

$\mathbb{R}^{3\times 3}$, i.e $R_B = [r_{B1}, r_{B2}, r_{B3}]$. $\mathcal{F}$ describes the respective legs and also the thrusters over it. The position of the foot can then be defined using forward kinematics equations shown.

$$\begin{aligned}
\boldsymbol{p}_{Fi} &= \boldsymbol{p}_B + R_B \boldsymbol{l}_{hi}^B + R_B \boldsymbol{l}_{fi}^B \\
\boldsymbol{l}_{fi}^B &= R_y(\phi_i) R_x(\gamma_i) \begin{bmatrix} 0, & 0, & -l_i \end{bmatrix}^\top
\end{aligned} \quad (3.2)$$

The position of the thrusters are defined as $\boldsymbol{p}_{ti}$ with respect to the body. The superscript $B$ represents a vector defined in the body frame, and the rotation matrix $R_B$ represents the rotation of a vector from the body frame to the inertial frame. Since the legs are considered massless, the kinetic and the potential energies of the HROM can be calculated with the equations shown below.

$$\begin{aligned}
K &= \left( \frac{1}{2} \dot{\boldsymbol{p}}_B m_B \dot{\boldsymbol{p}}_B^\top + \boldsymbol{\omega}_B I_B \dot{\boldsymbol{\omega}}_B^\top \right) \\
V &= -m_B \boldsymbol{p}_B^\top \boldsymbol{g} \\
L &= K - V,
\end{aligned} \quad (3.3)$$

where $\boldsymbol{\omega}_B$ is the body angular velocity in the body frame and $\boldsymbol{g}$ is the gravitational acceleration vector. Then the Lagrangian of the system can be calculated as $L = K - V$ and the dynamic equation of motion can be derived using the Euler-Lagrangian Method. In order to avoid the possibility of arriving at singularities through the simulation, the body orientation is defined using the Hamiltonian's principles and the modified Lagrangian for rotation in SO(3). Then the equation of motion can be derived as follows:

$$\frac{d}{dt}\frac{\partial L}{\partial \boldsymbol{\omega}_B^B} + \boldsymbol{\omega}_B^B \times \frac{\partial L}{\partial \boldsymbol{\omega}_B^B} + \Sigma_{j=1}^3 \boldsymbol{r}_{Bj} \times \frac{\partial L}{\partial \boldsymbol{r}_{Bj}} = \boldsymbol{u}_1, \qquad \frac{d}{dt} R_B = R_B [\boldsymbol{\omega}_B^B]_\times$$
$$\frac{d}{dt}\frac{\partial L}{\partial \dot{\boldsymbol{q}}} - \frac{\partial L}{\partial \boldsymbol{q}} = \boldsymbol{u}_2, \qquad (3.4)$$

where $\boldsymbol{u}_1$ and $\boldsymbol{u}_2$ are the sum of all generalized torques and forces respectively $[\cdot]_\times$ is the skew operator. The dynamic system acceleration can then be solved to obtain the into the following standard form [13]:

$$D_d(\boldsymbol{q}_d)\ddot{\boldsymbol{q}}_d + C_d(\boldsymbol{q}_d, \dot{\boldsymbol{q}}_d)\dot{\boldsymbol{q}}_d + G_d(\boldsymbol{q}_d) = \Sigma_{i \in \mathcal{F}} [B_{gi}\boldsymbol{u}_{gi}] + \boldsymbol{u}_e$$
$$B_{gi} = \frac{\partial \dot{\boldsymbol{p}}_{f,i}}{\partial \dot{\boldsymbol{q}}_d} \qquad (3.5)$$

where $\boldsymbol{q}_d = [\boldsymbol{q}_B^\top, \boldsymbol{q}_L^\top]$, $D_d$ is the mass/inertia matrix, $C_d$ contains the Coriolis vectors and gravitational vectors are defined in $G_d$, and $B_{gi}\boldsymbol{u}_{gi}$ represent the generalized force due to the GRF ($\boldsymbol{u}_{gi}$) acting on the foot $i$. The term $\boldsymbol{u}_e \in \mathbb{R}^6$ represents the external wrench acting on the rigid body of the HROM i.e $u_e = [f^\top, m^\top]^\top$. This represents the thrust force that acts on the Husky robot and condenses it into a wrench. The system is driven by setting the joint variable's accelerations to track a desired joint states. The joint states and the inputs are defined as follows

$$\boldsymbol{q}_L = [\boldsymbol{\phi}^\top, \boldsymbol{\gamma}^\top, \boldsymbol{r}^\top]^\top, \qquad \ddot{\boldsymbol{q}}_L = \boldsymbol{u}_L, \tag{3.6}$$

where $\boldsymbol{\phi}$, $\boldsymbol{\gamma}$, and $\boldsymbol{r}$ contain the joint variables of all legs (hip frontal, hip sagittal, and leg length respectively), and $\boldsymbol{u}_L$ forms the control input to the system in the form of the leg joint state accelerations. The full system of equations can then be derived from equation 3.5 and equation 3.6 as follows:

$$\begin{aligned}
\dot{\boldsymbol{x}} &= \begin{bmatrix} \dot{q}_s \\ D_s^{-1}(-C_d - G_d + B_d \boldsymbol{u}) \end{bmatrix} = \boldsymbol{f}(\boldsymbol{x}, \boldsymbol{u}), \\
&\boldsymbol{f} : \mathbb{R}^{n \times m} \mapsto \mathbb{R}^n \\
\boldsymbol{x} &= [q_d^\top, \dot{q}_d^\top], \boldsymbol{u} = [\boldsymbol{u}_e, \boldsymbol{u}_L, \boldsymbol{u}_g] \\
\boldsymbol{y} &= \boldsymbol{g}(\boldsymbol{x}, \boldsymbol{u}), \\
&\boldsymbol{g} : \mathbb{R}^{n \times m} \mapsto \mathbb{R}^p
\end{aligned} \tag{3.7}$$

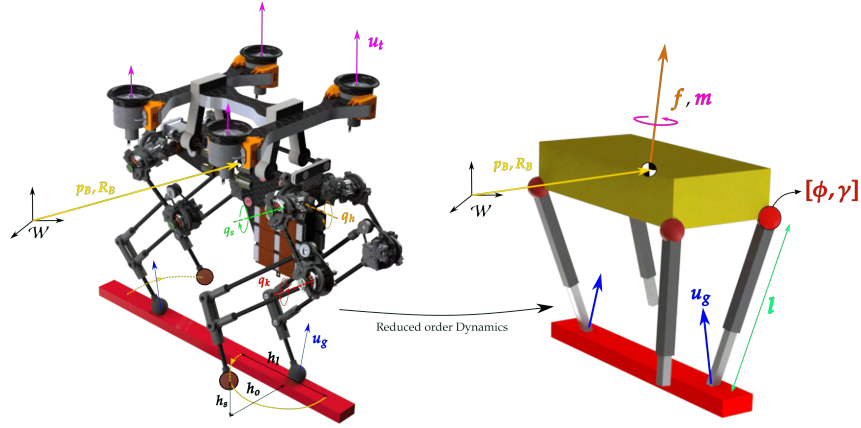

Figure 3.2: Modelling of the Husky Narrow path walk problem in the HROM model with the corresponding parameters that is translated

$\boldsymbol{x}$ is obtained by combining both the dynamic and massless leg states and their derivatives form the full system states. $\boldsymbol{u}$ is a vector of the inputs, and $\boldsymbol{y}$ is the outputs of the systems and is mapped from the states and inputs through $\boldsymbol{g}$. The GRF is modeled using a compliant ground model and Stribeck friction model, defined as follows:

$$\begin{aligned}
\boldsymbol{u}_{gi} &= \begin{cases} 0 & \text{if } z_i > 0 \\ [u_{gi,x}, u_{gi,y}, u_{gi,z}]^\top & \text{else} \end{cases} \\
u_{gi,z} &= -k_{gz} z_i - k_{dz} \dot{z}_i \\
u_{gi,x} &= -s_{i,x} u_{gi,z} \operatorname{sgn}(\dot{x}_i) - \mu_v \dot{x}_i \\
s_{i,x} &= \left( \mu_c - (\mu_c - \mu_s) \exp\left(-|\dot{x}_i|^2 / v_s^2\right) \right),
\end{aligned} \tag{3.8}$$

where $x_i$ and $z_i$ are the $x$ and $z$ positions of foot $i$ respectively, $k_{gz}$ and $k_{dz}$ are the spring and damping coefficients of the compliant surface model respectively, $u_{gi,x}$ and $u_{gi,y}$ are the ground friction forces respectively, $\mu_c$, $\mu_s$, and $\mu_v$ are the Coulomb, static, and viscous friction coefficients respectively, and $v_s > 0$ is the Stribeck velocity. The derivations of $u_{g,y}$ follow similarly to the derivations to $u_{g,x}$.

# Chapter 4

# Control Design

In this chapter, I elaborate on the structure of the Simulink ®Realtime firmware to control the Husky robot. Also, I outline the open loop trajectory generation of walking gaits. To reach this end, I employ a method to schedule gaits using Bézier control points. The generated walking gait allows the robot to move forward on the ground. To stabilise the frontal dynamics and reduce any roll error, the Husky is modelled as a linear inverted pendulum in the frontal plane. To compensate for this roll error, thruster forces are generated to on either side. This work is built on previous work done on Husky which included a event-based open loop gait generator along with a stable Inverse Kinematics block [31].

## 4.1 Operating Husky using Matlab®Simulink

The communication pipeline of Husky was explained in chapter 2. The host computer runs Matlab® 2019b and communicates with the Speedgoat Realtime machine through EtherCAT. The Simulink® code is compiled on the host PC generating a C code and then flashed on the Speedgoat Realtime machine. An EtherCAT communication protocol is initiated between the host computer and the Speedgoat realtime machine. This way we can send commands to the robot in the form of 3D Bézier control points and user-inputs and receive data in the form of Encoder data and other information such as error codes, and control words. The neutral foot position co-ordinates are also user-inputs. This is because the joint sensors are relative encoders and an absolute reference in the form of a neutral position. The control points are generated as a function of the user inputs such as the step length, step height, number of steps, and the trajectory points are generated as a function of the gait time and Bézier trajectory function defined in section 4.2.

A sequence generator block regenerates the user-inputs based on the number of steps and the number of loops. A gait-scheduling concept makes it possible to make the firmware gait-agnostic and is explained in subsection 4.2.1. A spatial illustration of this desired trajectory is shown in Fig. 4.2 along with the frontal and sagittal planes. The target foot location is sent to an Inverse Kinematic block that calculates the required joint angles. A saturation block is used to limit the joints from moving to points in the configuration space that can cause self collisions and a rate-limiter block is used to prevent the joints to move egregiously fast.

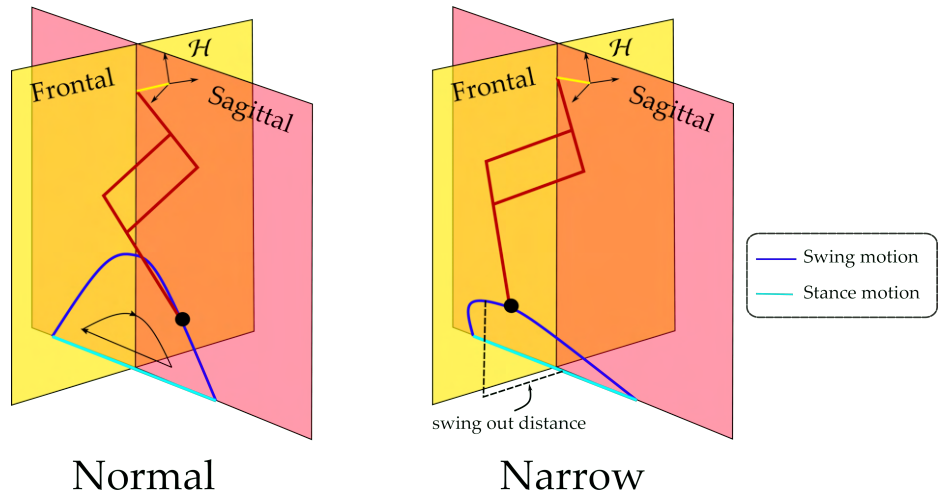

Figure 4.2: Spatial Illustration of the generated trajectory while also depicting the frontal, sagittal planes along with the position of the hip frame $\mathcal{H}$.

## 4.2 Bézier polynomials

The Bézier polynomial is employed in generating parametric curves by utilizing Bézier Control Points. The trajectory of the polynomial is determined through a timing parameter that ranges from 0 to 1. This timing parameter is updated in each iteration based on the looptime. Bézier polynomials, with their control points, offer the advantage of ensuring continuous and smooth trajectories, and guarantees that the path always forms a convex hull of the control points. A Bézier polynomial of degree M with $b_i$ control points parametrized over s (gait timing variable) ranging from 0 to 1 is defined by M + 1 coefficients, $\alpha_k^i$ per i-th control variable [1]:

$$b_i(s) = \sum_{k=0}^{M} \alpha_k^i \binom{k}{i} s^k (1-s)^{(M-k)}$$

$$\text{where, } \binom{k}{i} = \frac{k!}{i!(k-i)!},$$

(4.1)

### 4.2.1 Gait scheduling

To achieve various walking gaits, specifically the two-point contact and three-point contact walking, we employ a gait scheduling method. This method enables the submission of a block of Bézier control points as a user input, facilitating the generation of periodic gaits. An overview of this gait scheduling method is depicted in Fig. 4.1. The motion of each leg can be described using a set of n-spatial Bezier control points, forming a $3 \times n$ matrix. By combining these blocks from all four legs, we obtain a $12 \times n$ matrix representing the complete trajectory block for all legs within a specified duration. This way multiple trajectories can be stacked on top of each other as a sequence block of size $12k \times n$ and can be repeated multiple times specified by the user.

For example, to perform one gait, consisting of a step and stance, we design it as two separate trajectories as the 2 pairs of legs are completely out of phase with each other. Each of these trajectories are made of $n = 5$ Bézier control points. This makes the input trajectory block of size $24 \times 5$. This is the function of the controller manager in the Fig. 4.1. The sequence manager structures this trajectory block and creates a decision process of moving to the next trajectory depending on the progress of the ongoing trajectory and also keeps a count of the number of times a sequence based on the user input. To facilitate this, the current foot position is always kept as a memory. Based on the sequence, the current foot position and the progress of the gait, the sequence manager sends the target Bézier control points and the Bézier timing parameter $s$. Based on the equation, 4.1 the gait generator gives out a vector of length 12 that has the target position of each foot in Cartesian co-ordinates. These cartesian co-ordinates are then sent to the IK function to calculate the target joint angles.

This flexible structure of trajectory generation has made it such that the input only needs a properly structured matrix of Bézier control points. This structure also can be used when the trajectory is no longer a user input and is an output of a controller.

## 4.3  MPC using locally linearized Dynamics

In this section, I present a linearized MPC controller that can be used to control the pose of the HROM model using linearized dynamics. This method has shown that under reasonable

We use thruster inputs, modelled as a wrench, as an optimal control input by minimizing a cost function over a predictive horizon. In chapter 3 I derived the dynamics of the HROM as $\dot{x} = f(x, u)$ and the output of the system as $y = g(x, u)$. This is a highly non-linear function and implementing a inexpensive optimal controller using this model can become very tedious, expensive and is outside the scope of my thesis research. The linearized state space form, along with the A,B,C and D, matrices from the non-linear state space equations can be obtained by equating the jacobians as follows,

$$\begin{aligned} A &= \tfrac{\partial f}{\partial x}|_{x_0}, B_e = \tfrac{\partial f}{\partial u_e}|_{x_0} \\ B_j &= \tfrac{\partial f}{\partial u_j}|_{x_0} \\ C &= \tfrac{\partial g}{\partial x}|_{x_0} \end{aligned} \quad (4.2)$$

To calculate the aforementioned Jacobians quickly, a numerical strategy can be employed where a sufficiently small perturbation is applied to each element of the input and corresponding output is measured. The complete numerical formula to calculate the Jacobians can then be written as follows,

$$\begin{aligned} A_{n \times n} &= [\ldots, \frac{\partial \boldsymbol{f}}{\partial x_{i-1}}, \frac{\partial \boldsymbol{f}}{\partial x_i}, \frac{\partial \boldsymbol{f}}{\partial x_{i+1}} \ldots] \\ \frac{\partial \boldsymbol{f}}{\partial x_i} &= \frac{[\boldsymbol{f}(\ldots, x_i + h, \ldots) - \boldsymbol{f}(\ldots, x_i - h, \ldots)]}{2h} \end{aligned} \quad (4.3)$$

where $h$ is a perturbation constant that is in the order of $10^{-5}$. A similar method can then be used to find the B matrices as well. Consequently, the linearized dynamics at the operating point can then be written as,

$$\begin{aligned} \dot{\boldsymbol{x}} &= A\boldsymbol{x} + B\boldsymbol{u} \\ \text{where,} \quad B &= [B_e, B_j], \end{aligned} \quad (4.4)$$

where B is a block matrix consisting of the $B_e$ and $B_j$ calculated as shown in equation 4.2 and equation 4.4. Using an Explicit-Euler methodology, a discretized version of this linearized dynamics can be formulated as follows,

$$\begin{aligned} \boldsymbol{x}_{k+1} &= \boldsymbol{x}_k + A_d \boldsymbol{x}_k + B_d \boldsymbol{u}_k \\ \text{where,} \quad A_d &= A \times dt, B_d = B \times dt, \end{aligned} \quad (4.5)$$

$dt$ is the control loop time. This linearized dynamics is marched forward for N time steps and the cost function is calculated for the N-horizon. Finally, the optimal control input to the body becomes a solution of the optimization problem,

$$\operatorname*{argmin}_{\boldsymbol{u}_e} \quad \sum_{k=1}^{N} \boldsymbol{x}_{e,k}^{\top} Q \boldsymbol{x}_{e,k} + \boldsymbol{u}_{e,k}^{\top} R \boldsymbol{u}_{e,k}$$
$$s.t \quad \boldsymbol{x_e} = \boldsymbol{x}_{k,ref} - \boldsymbol{x}_k,$$
$$\boldsymbol{u}_{e,min} < \boldsymbol{u}_e < \boldsymbol{u}_{e,max}, \quad (4.6)$$
$$\boldsymbol{x}_{k+1} = \boldsymbol{x}_k + A_d \boldsymbol{x}_k + B_d \boldsymbol{u}_k,$$

$\boldsymbol{x}_e \in \mathbb{R}^3$ is an error term that calculates the error of the body pose in the form of Euler angles calculated from the body transformation matrix $\boldsymbol{R_B}$. $Q \in \mathbb{R}$ and $R$ are cost function weights of their corresponding terms. The first optimal control input for the horizon is then used as the control input for the HROM and is held until the dynamics is linearized again in the future with new optimal control inputs.

## 4.4 MATLAB®Simulation

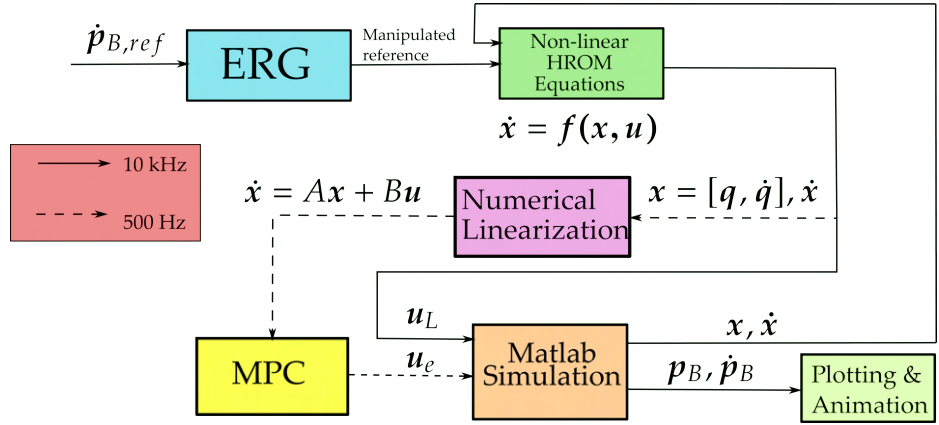

Figure 4.3: Flowchart depicting the control schematic and data flow in the Matlab®simulation. ERG [28] [8](Explicit Reference governor), manipulates the reference applied to make sure the leg doesn't break friction cone constraints

The MATLAB®simulation was implemented on the previously defined HROM [28] [45] on which ERG (Explicit Reference Governor) was implemented to manipulate the body forward velocity reference to satisfy ground friction cone constraints. The stance and swing legs switch based on a timing based state-machine, where the gait time is defined as an user-defined parameter. In a way, the ERG is a controller that indirectly controls the joint angles accelerations. The control input for the external wrench is calculated as an optimal solution of the MPC problem as proposed in section 4.3. The full vector of control inputs along with the derivatives of the states is marched forward using a 4th order Runge-Kutte (RK4) numerical solution to obtain the next states given the current states and it's derivatives, along with the inputs. The RK4 integration scheme is formulates as follows,

$$\begin{aligned} \text{Given,} \quad & \dot{\boldsymbol{x}} = \boldsymbol{f}(\boldsymbol{x}, \boldsymbol{u}) \\ & \boldsymbol{f1} = \boldsymbol{f}(\boldsymbol{x}, \boldsymbol{u}) \\ & \boldsymbol{f2} = \boldsymbol{f}(\boldsymbol{x} + \boldsymbol{f1}\frac{\delta t}{2}, \boldsymbol{u}) \\ & \boldsymbol{f3} = \boldsymbol{f}(\boldsymbol{x} + \boldsymbol{f2}\frac{\delta t}{2}, \boldsymbol{u}) \\ & \boldsymbol{f4} = \boldsymbol{f}(\boldsymbol{x} + \boldsymbol{f3}\delta t, \boldsymbol{u}) \\ & \boldsymbol{x}_{k+1} = \boldsymbol{x}_k + (\frac{\boldsymbol{f1}}{6} + \frac{\boldsymbol{f2}}{3} + \frac{\boldsymbol{f3}}{3} + \frac{\boldsymbol{f4}}{6})\delta t \end{aligned} \quad (4.7)$$

where $\delta t$ is the loop time at which the simulation is run.

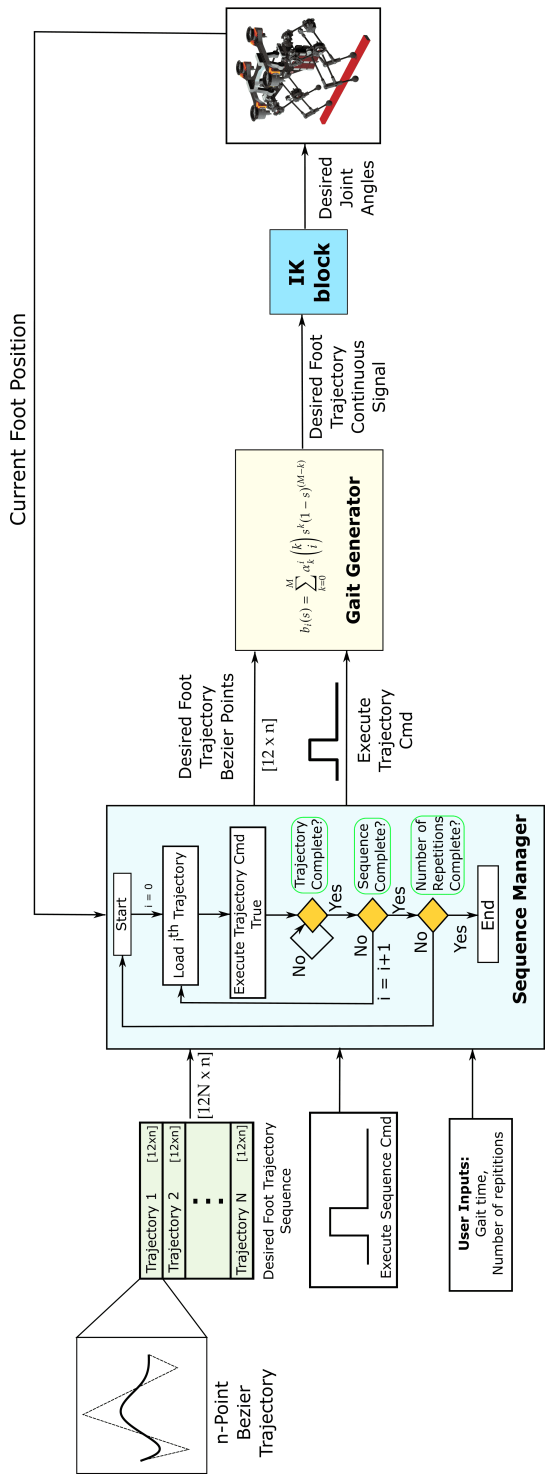

Figure 4.1: Structure of the Simulink®firmware that exhibits the gait scheduling strategy

# Chapter 5

# Simscape Simulation Results

In order to test and validate the gaits of the robot generated, along with the working of the inverse kinematics block, I built a high-fidelity simulation of Husky on Simscape multibody. This allows us to avoid self collisions, monitor joint torques and get an estimate of the joint torques. It also help us understand the scale of the thruster forces on each EDF. To do this, we need to provide Simscape with a URDF (Universal Robot Descriptor Format) generated from the Solidworks model of the robot. As a result, we test the expected behaviour of the system in a physics environment and also have room for further development to build controllers on this high fidelity model.

## 5.1 SimScape Husky Model

The Solidworks-generated URDF model of Husky was integrated into the Simulink® environment using the 'smimport' functionality. Within the URDF file, the mass, inertia, joint configurations, and actuation details of the components were specified. Trajectory inputs for the joints were generated based on the desired gait, employing the IK block as a MATLAB®  function within the simulation and sent to the hip, sagittal, and knee joints, all defined as revolute joints. The corresponding torques required for joint actuation were automatically computed by the MATLAB® solver. To translate Simulink®  signals into physical commands for joint positioning, the Simulink-PS converter was employed. Additionally, transform sensors were utilized to ascertain the pose and velocity of the body relative to a global world frame.

The URDF file is complemented by STL meshes for individual parts, enhancing the robot's visualization. Figure 5.1 illustrates the Simscape model of the robot within the mechanics explorer. Leveraging these meshes, a convex hull of the parts is generated, and a contact model is established between each part and the ground. User-defined parameters for coefficients of friction and ground stiffness contribute to this contact model. The ground contact block doubles as a sensor, calculating ground reaction forces upon foot-ground contact, facilitating verification of friction cone constraint adherence. Convex hull geometries of the body are also derived to inspect potential contact points between the legs and body. Furthermore, supplementary rigid blocks attached to the body (shown in orange in Fig. 5.2), simulate the position of the power electronics rack on the robot, with meticulous consideration to prevent inadvertent contact.

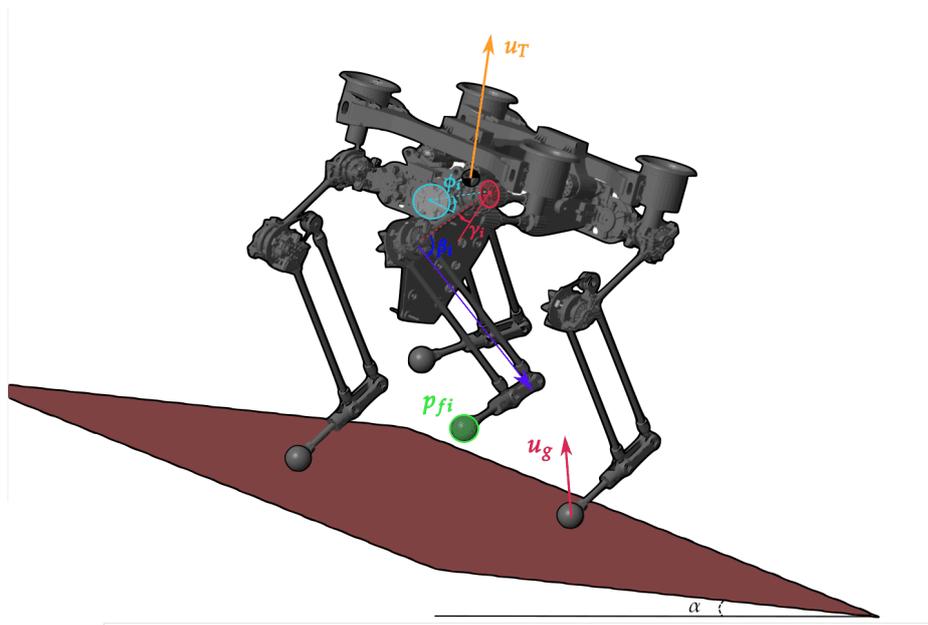

Figure 5.1: Depicts the visualization of the URDF model and Simscape environment developed to realistically evaluate Husky's gait performance.

To activate the thrusters, a force and moment block is positioned at the center of each thruster. In the physical robot, these thrusters play a crucial role in stabilizing the robot's pose through the Pixhawk controller. In the simulation, to compensate for the error in the frontal plane (i.e., roll error), the robot is modeled as a simple inverted pendulum, while the thruster compensation is represented by a simple PID controller, as depicted in Figure 5.2. Simulation results for foot normal forces, joint angles, and thruster forces are further explained in Section 5.2. Overall, the structure of the Simscape multibody model is designed to mimic the actual Simulink®  Realtime firmware along with the trajectory generation pipeline. This allows for a comparison of outputs between the simulation and the robot, aiming to bridge any observed gaps.

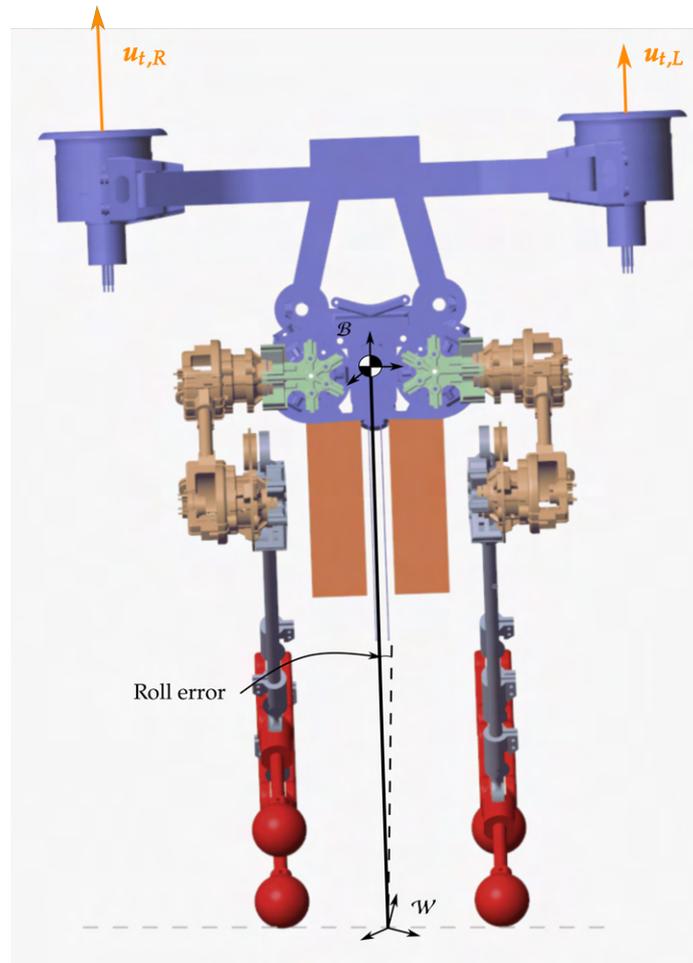

Figure 5.2: Depicts how roll error is defined in the frontal plane and thruster forces to compensate for this

## 5.2 Simulation Results

To stay within the scope of the thesis, the robot was constrained in the pitch and yaw degree of freedoms. The simulation settings were set to the ode45 Runge Kutta solver with a variable time step. For the sake of visualization a path of width 1m is simulated in the floor The simulation of thruster-assisted walking was run for a duration of ten seconds, and the robot moved forward 1.6 m. A snapshot image Fig. 5.3 shows the motion of the robot. The trajectory of the legs was generated with a step length of 0.1 m and step height of 0.1 m. The robot is successfully able to walk without any self-collisions and the roll error is bounded, as seen in 5.6a. The roll compensation is possible due to the thruster actuation shown in Fig. 5.5. From the thruster actuation it can be observed that during two point contact walking, the robot swings from side to side and the average thruster forces from either side generates a torque and corrects for the robot's roll.

Foot normal forces as shown in Fig. 5.7a depicts a typical transient-impulse behaviour, commonly seen in most walking systems. Due to the open loop behaviour of the walking system, for a short period of time we can notice that some of legs bounce off the ground due to the impulse force generated, and there is an absence in ground reaction force. For the simulations performed, a standard stick-slip friction model is used and a static friction coefficient of 0.5 and dynamic friction coefficient of 0.3 was used which are reasonable values for considering a plastic to concrete contact. Considering the point foot contact in the simulation, some slippage during the stance phase was also evident, which shows that the friction cone constraints are not satisfied during those points and the legs slip. This observation shows that some form of closed loop control is necessary to manipulate the ground reaction forces to resist slippage.

Similarly a narrow walking gait was also simulated and a snapshot is shown in Fig. 5.4. Here the width of the ground visualization is about 0.1 meters. The step length is set as 0.08 m and the step height is 0.1 m. For the legs to avoid colliding with other, the leg swings outwards with respect to the body during the swing phase. The 'swing out' distance here is set to be 0.07m and the gait period was set to be 0.5 secs. For the following open-loop gait parameters, the robot is able to balance itself in the frontal plane with the help of the thrusters and also be able to move forward.

The joint trajectories for the normal and narrow walking gaits are shown in Fig. 5.8. In the normal walking gait, the frontal motor barely moves, only in the order of 0.003 degs, and the leg only moves in the saggital plane. The sagittal angle sweeps an angle of 10 degrees during every gait and the knee sweeps an angle of 16 degrees. During, narrow path walking we see the hip frontal angle also sweep an angle of 4 degrees. This is because of the leg has to swing outside to avoid collisions

These simulation results also helped me gain insight into the behaviour of the physical system and the fact that the robot does not have any self collisions. It also helped me choose the optimal values for running experiments on the physical platform.

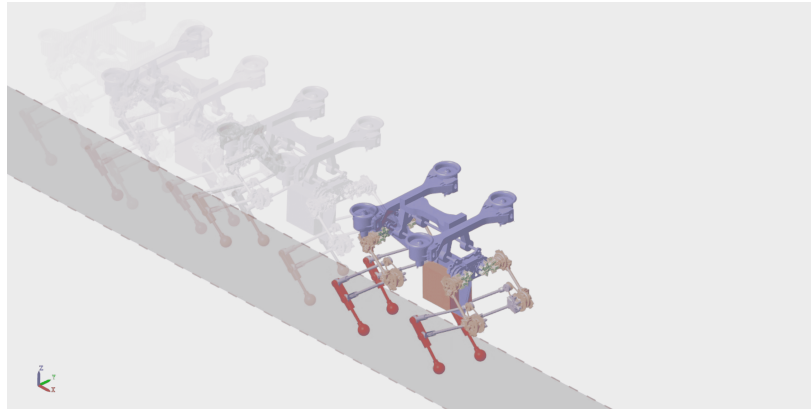

Figure 5.3: Shows snapshots of Husky walking in the Simscape environment using a two-point contact gait along with thruster assistance

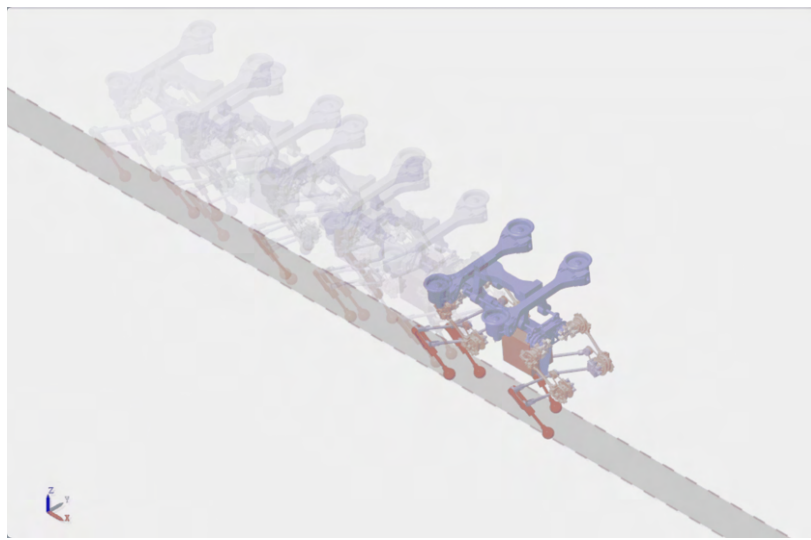

Figure 5.4: Snapshots of Husky walking on narrow platform in simulation

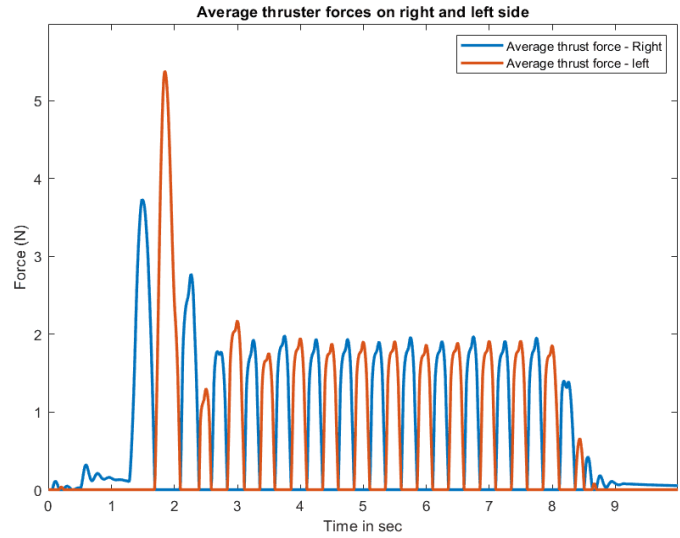

Figure 5.5: Depicts the visualization of the URDF model and SimScape environment developed to realistically evaluate Husky's gait performance.

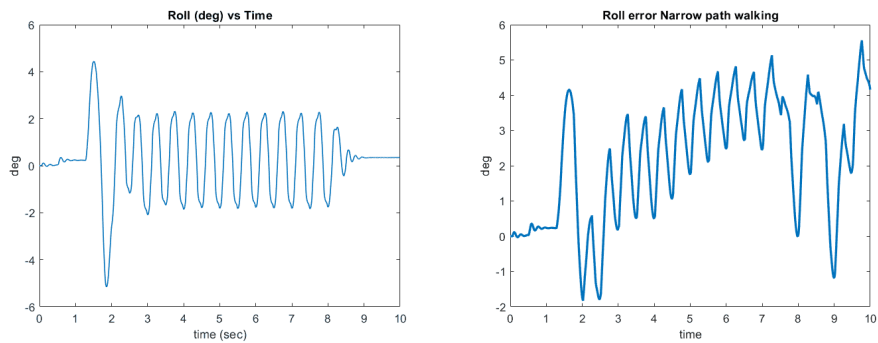

(a) Roll error normal thruster assisted walking
(b) Roll error during narrow path walking

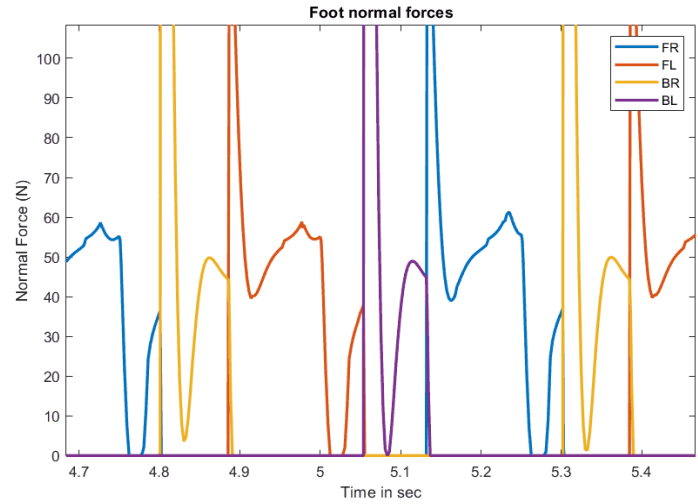

(a)

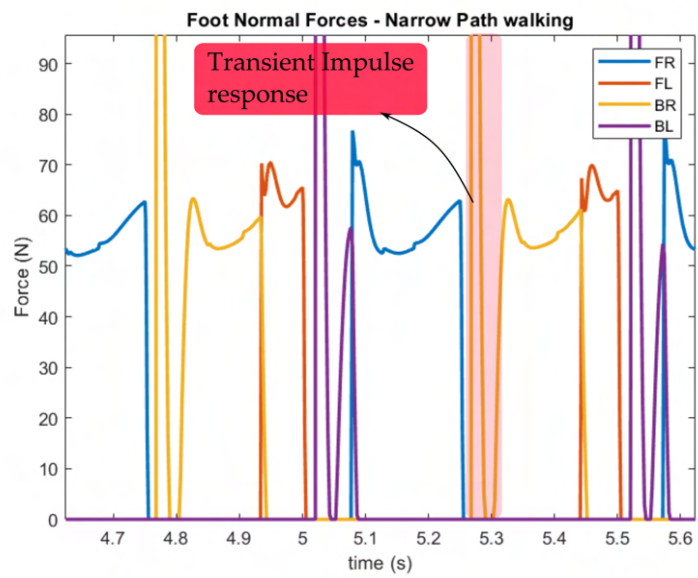

(b)

Figure 5.7: Plot of Foot normal forces in (a) normal walking and (b) narrow walking showing the transient impulse response when the feet strike the ground and a short period when the foot lifts of the ground

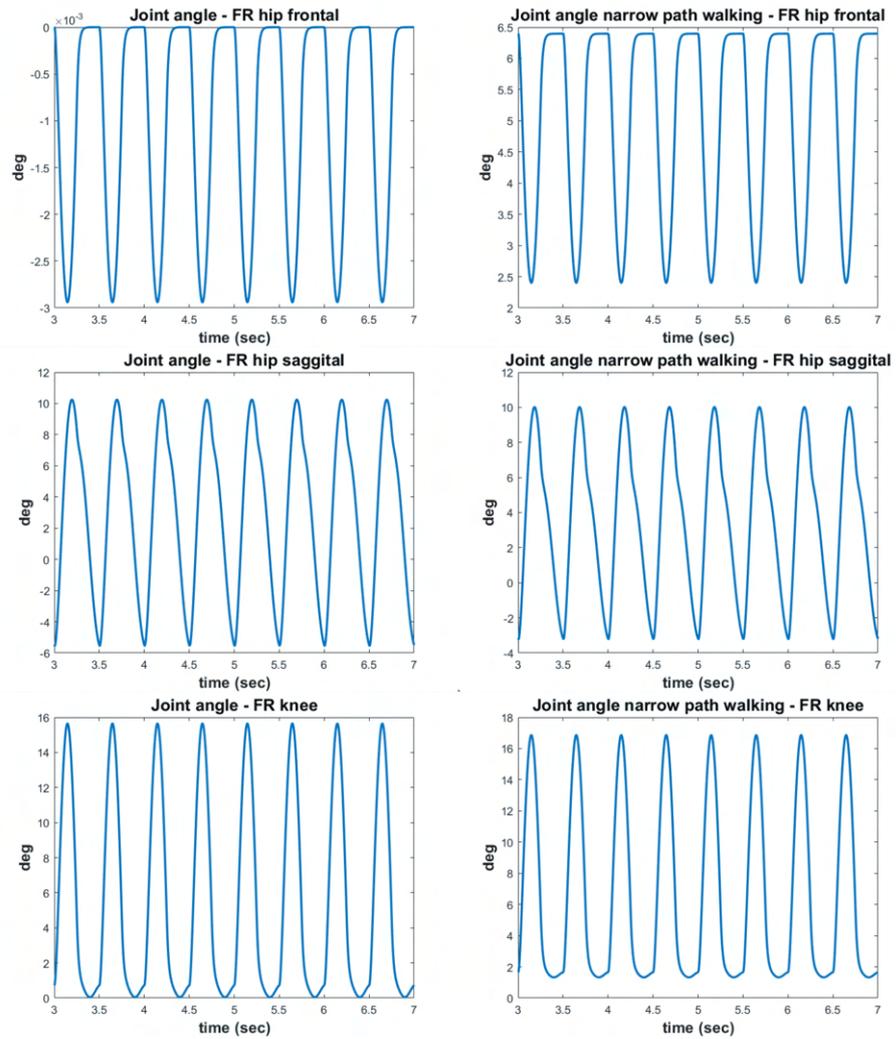

Figure 5.8: Joint trajectories of the robot during normal and narrow path walking

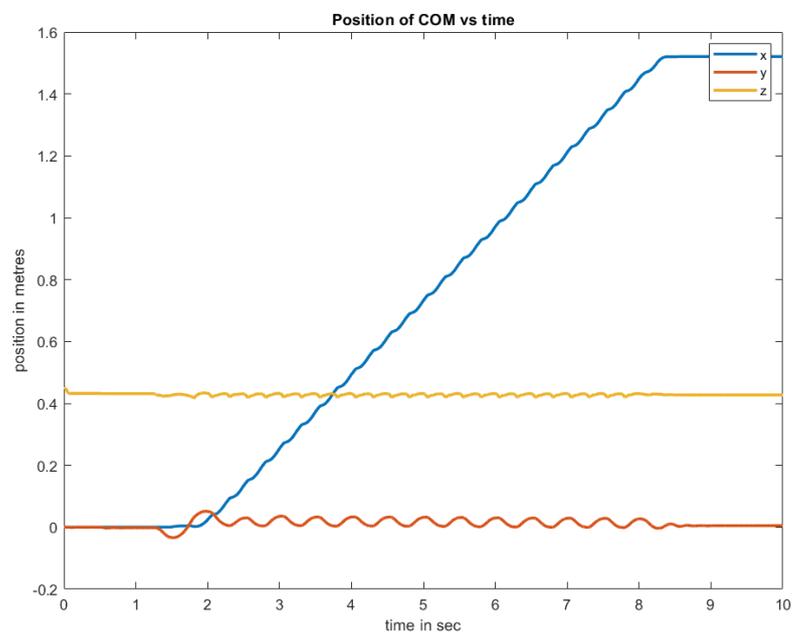

Figure 5.9: Plot of the COM position of the robot against time

# Chapter 6

# Experimental setup and Results

In this chapter, I discuss the experimental results performed in *SiliconSynapse* Labs at Northeastern University. I describe the test platform used for the performing these tests: :namely a thruster assisted trotting and the first walking test on Husky. I then further discuss the test outcomes and outline any takeaways.

## 6.1 Husky Test Arena

The purpose of the Husky test setup, hereby referred to as gantry, shown in Fig. 6.1, is to provide an isolated test setup with tethered support to carry and lift the robot using ropes and an external power supply to power the joint actuators. The gantry is built using 1" aluminum extrusions which also house cameras to record experiments taking place. The gantry is also fitted with wheels to help continuously support and power the robot if it moves. An E-stop button is present on the gantry to kill the power electronics in case of any emergency. The Speedgoat RT machine is also mounted on the gantry and is connected with the host PC. To power the thrusters, the frame is fitted with a stand to house the LiPo batteries. This setup is designed in such a manner test the functionality of Husky with repeatable outcomes.

The emergency stop unit on ELMO amplifiers is connected to the Safe Torque Off (STO) unit. When activated, this unit prevents the drive from generating any torque. The emergency stop button is conveniently located for quick access in case of unexpected events or user input errors. The external power supply is the Extech 382275 600W Switching Mode DC Power Supply. It allows adjustable output ranging from 0 to 30 Volts DC and 0 to 20 Amps, with a resolution of 0.1 Volts/Amps. For our specific application, we configured the output to 30V with a current limit set at 14A. The Speedgoat RT machine is currently powered separately through a power plug with a 12 V AC/DC adapter.

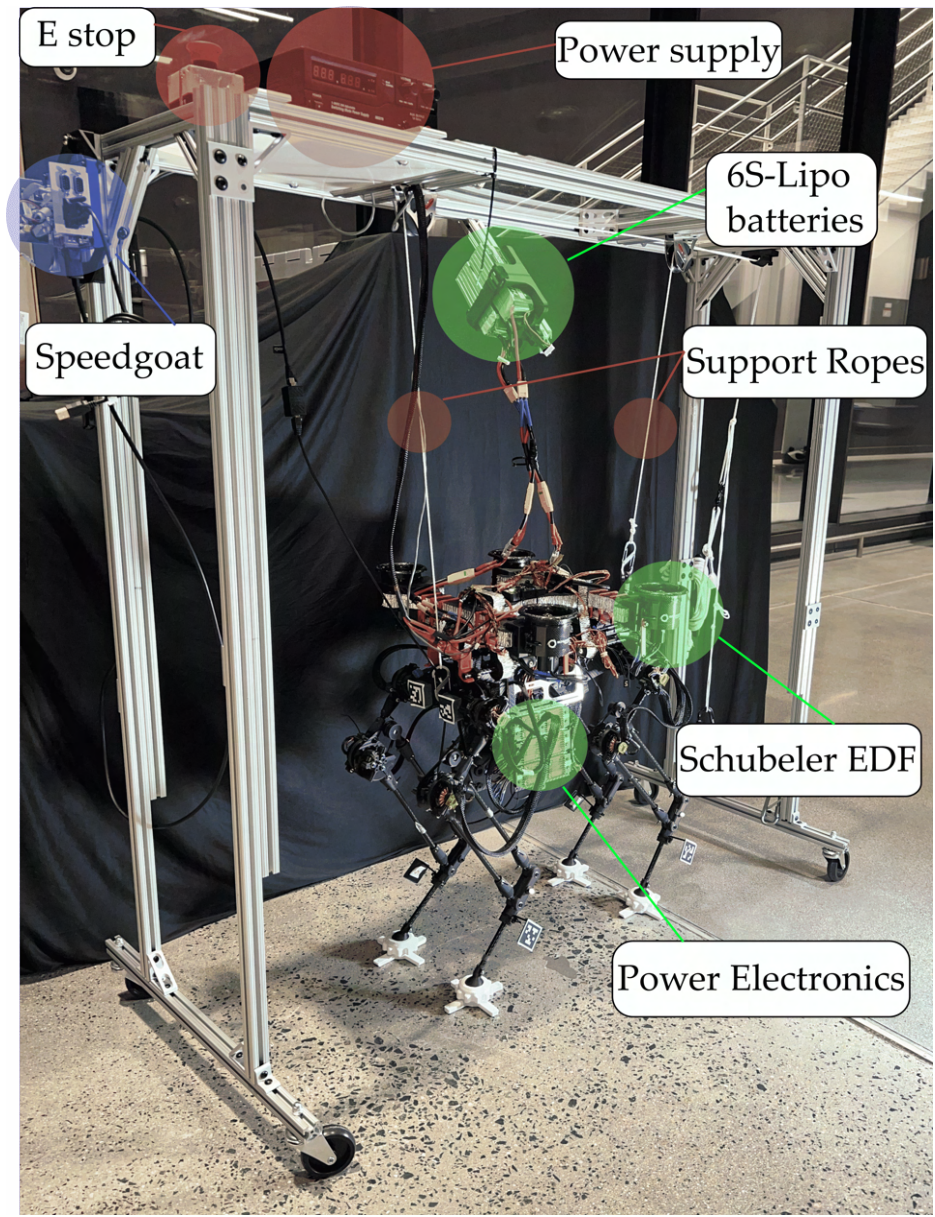

Figure 6.1: Shows movable Husky test gantry, at *SiliconSynapse* Lab in EXP Northeastern University. The robot stands 3ft tall and 1.5 ft wide

The communication pipeline of Husky was defined in Chapter 2. The host PC is running MATLAB® 2019b with the Simulink® realtime code developed to be compiled on the target. The Simulink® realtime code generates a C code which is deployed on the target and all communication is conducted via TCP/IP. During the testing, the robot's power electronics are switched on and the motors are enabled. With the help of the EtherCAT communication from the host PC, position commands for the actuator is given. Just as the test starts, the support rope takes the entire weight of the robot. As all the motors are enabled, the robot draws a total of 1.7 amps at 30 Volts. The robot is then slowly lowered from the ropes, with the motors enabled and commanded to hold position. With the robot lowering, the motors start taking a larger load and starts drawing current to hold the same position. The robot is lowered until a safe operating current is reached. Any experiment is carried with careful consideration of safety precautions and any untoward actuator behaviour is stopped using the E-stop button. The temperature of the amplifiers and the actuator windings are monitored using a infrared thermal camera and laser temperature gun. The support ropes also allow for testing the robot under different loading conditions.

For data collection purposes, The robot is fitted with Optitrack markers and the firmware is also capable to read positions from the Optitrack camera setup. Furthermore, the Optitrack setup can also be used as a position feedback strategy. To estimate the pose of the robot body, an onboard IMU calculates acceleration, gyroscope data and Euler angles in the body frame. The actuator encoders collect joint position data and track the joint positions. Simulation output data is saved on the RT machine and subsequently saved on the host PC for data collection and documentation.

## 6.2 Thruster-assisted trotting and walking

Owing to the slippage on the point feet in the Simscape walking simulation, Husky's feet was fitted with 3D printed cross feet to increase the support polygon and be able to test in a more stable and repeatable manner. The trotting gait was performed using the two point contact gait and the thruster forces controlled using a remote controlled joystick connected wirelessly to a Pixhawk and ESC's to provide the right amount of corrective thrust to the robot body. Three consecutive trotting tests were done and the The Pixhawk is set to 'Stabilize' so any control input from the user is modulated to make sure the Euler-angles of the body is stabilized. Data collected by the Pixhawk is stored in a microSD card. The voltage drop of the batteries during the thruster-assisted trotting

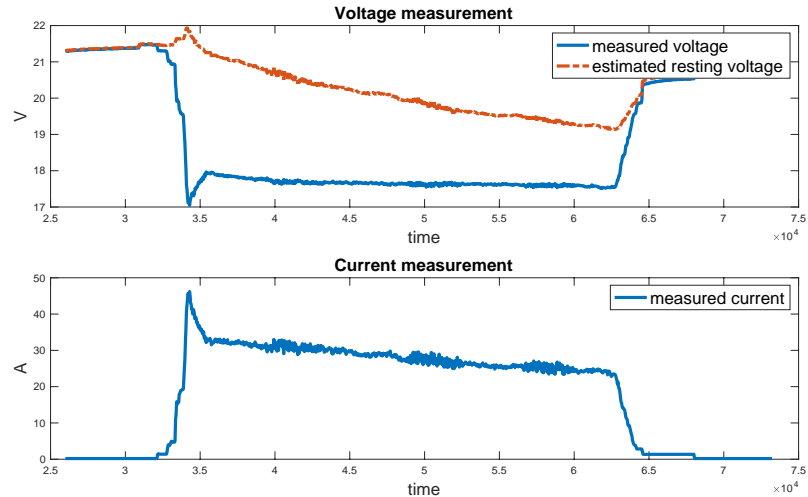

Figure 6.2: Depicts the voltage drop and current drawn by a pair of EDF's from information collected during a test performed in conjunction in research from [57]

experiment is plotted in Fig. 6.2 and the corresponding orientation is shown in Fig. 6.3. From observations, it can be seen that during trotting, the thrusters are able to stabilise the roll of the robot. The pitch and yaw of the robot has a bias and errors, because the robot is held with the support ropes and this restricts it's motion in pitch and yaw.

For thruster-assisted walking, we utilise a walking gait similar to the one seen in the simulation. We also test the ability of the gait generation system by stacking two identical trajectories. The generated trajectory for a single foot can be seen in Fig. 6.5 the corresponding joint trajectory from the IK block is shown in Fig. 6.6. The joint trajectories show same behaviour as compared to the simulations and also avoided any slef collisions. The generated trajectories is also smooth and continuous at the point the trajectory moves from the first block to the second. Implementing the trajectory on the robot showed us similar behaviour as per the Simscape simulation and the robot was able to walk forward. A snapshot of this walking test is shown in Fig. 6.4. Using optitrack and reflective markers attached to the leg, the 3D trajectory of the leg is obtained and is shown in Fig. 6.7. During the entire test, the robot draws 2.0 amps from the external power supply, while the support ropes took some weight off the robot reducing the load on the actuators.

These experiments provide us with promising early results and showcases the ability of a legged aerial system. The walking test helped us prove that such a lightweight structure still has the ability to generate stable walking gaits, while the thruster-assisted trotting results gave us further confidence about how the thrusters can influence the pose of the body while the robot performs legged maneuvers.

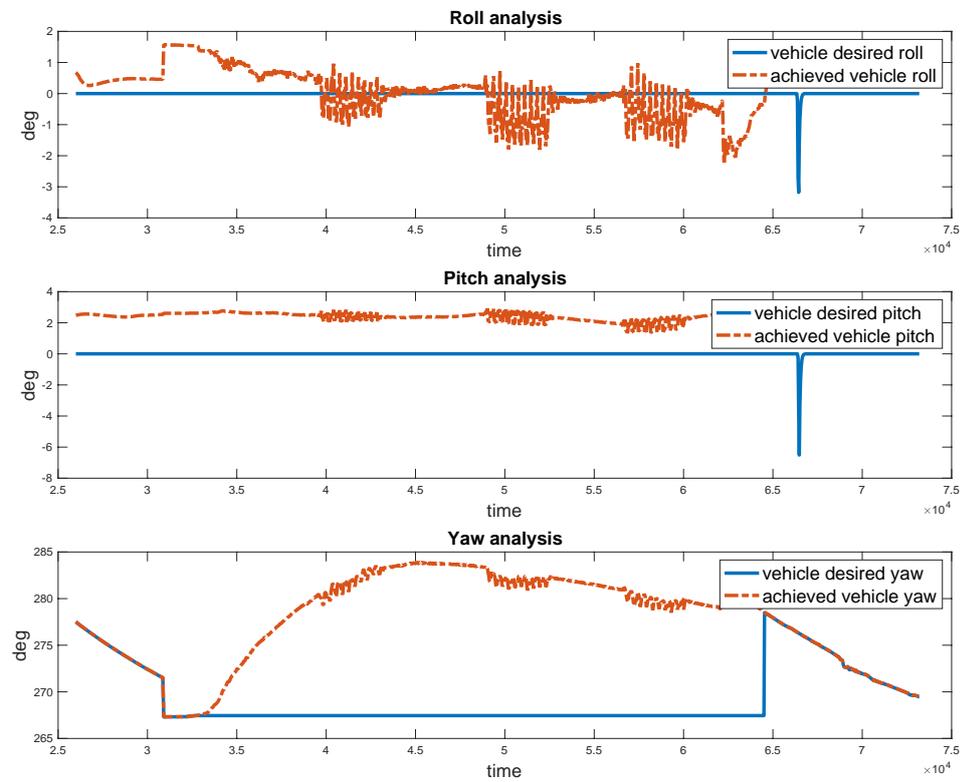

Figure 6.3: Depicts the orientation of the robot during thruster assisted trotting collected during a test performed in conjunction from [57]

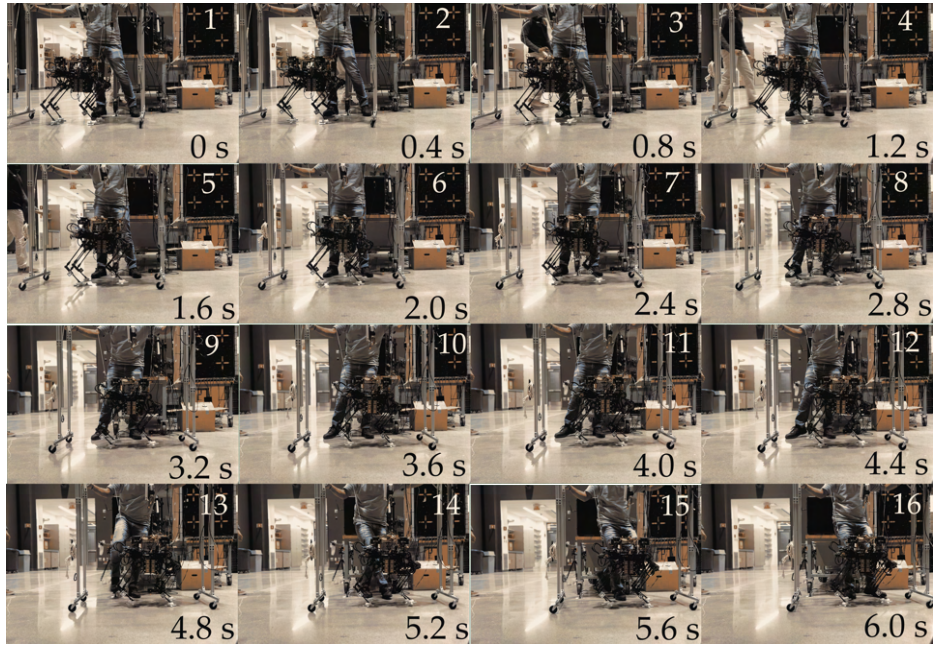

Figure 6.4: Snapshots from walking experiment, gait time = 0.5 secs, step length = 0.1m

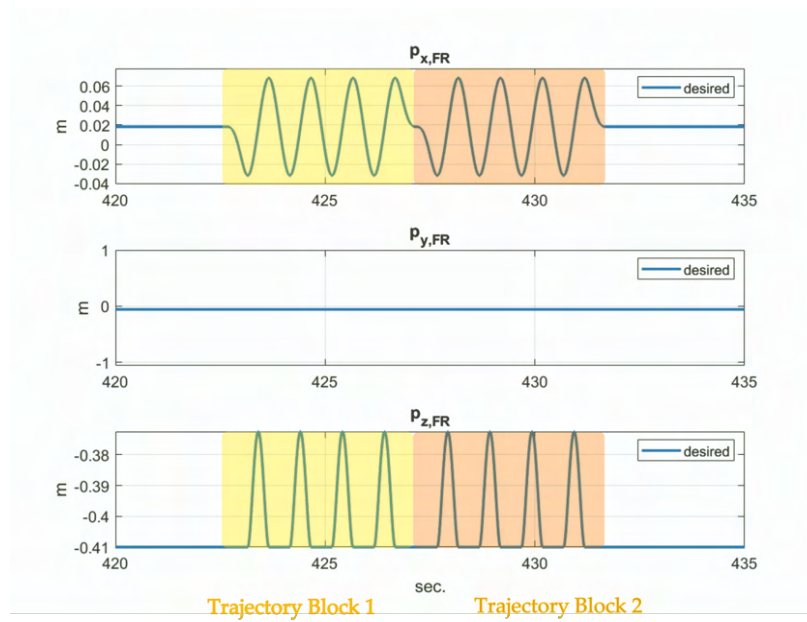

Figure 6.5: Desired trajectory of a single foot (FR), generated by repeating the same trajectory block showing a continuous transition while going from end of the trajectory to the start

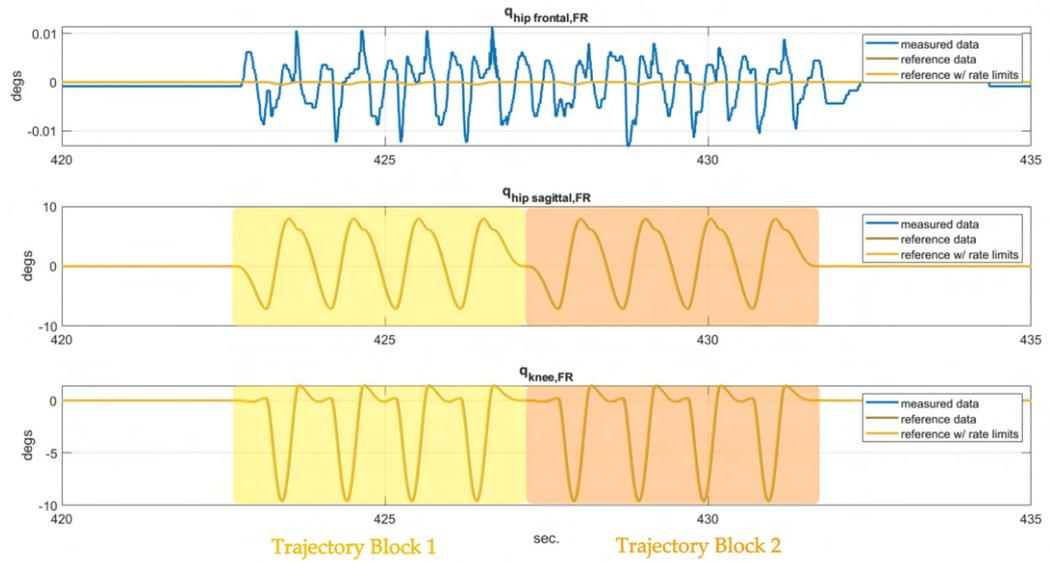

Figure 6.6: Joint angles of the FR leg during walking experiment

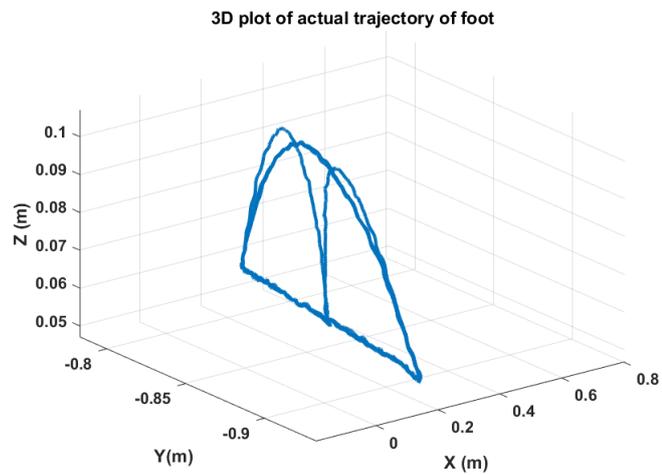

Figure 6.7: 3D-plot of the foot end trajectory obtained using optitrack, showing the swing and the stance motions along with the first half step the robot takes while walking and the last half step to align all the legs

# Chapter 7

# Conclusion

In this thesis, I propose the implementation of a sophisticated ability for legged robots to traverse narrow paths. This ability is inspired by observations of certain animals that demonstrate such behavior. The implementation is intended for Northeastern University's Husky, leveraging its multi-modal legged and aerial capabilities and can be a transferrable skill to be implemented in different scenarios.

The reduced order model developed previous research efforts [47] [28] has been used to implement a locally linearized MPC controller that calculates optimal external wrench inputs to stabilize the pose of the robot given fixed foothold location and open loop gaits. Simultaneously, the Simscape high fidelity model of the robot was also used to validate the gait generation methods used on the actual robot and ensure that any gait generated does not contribute to self collisions or unstable behaviours. The Simscape simulation also showed the ability of the robot to balance itself in the frontal plane using corrective thruster force action. Although the Simscape simulation has the ability to validate the gaits, to have a robust simulation of the gaits, I would like to implement a torque controller for each joint and consequently be able to implement the same MPC on the physical robot. The outcome of this thesis then defines the legged locomotion capabilities of Husky Carbon for a walking and thruster-asssisted trotting. The initial experimental results was done using a Pixhawk, equipped with it's own IMU, that controls the thrusters just to keep the propulsion unit roll, pitch, and yaw stable. An implementation of the proposed optimal controller using the HROM is required to stabilize the robot in a closed loop fashion.

Further work on the controller itself includes the ability to tightly couple the dynamics between the legged and aerial system and optimally using the ground reaction forces as well as the thruster inputs to balance the robot. Furthermore a higher fidelity on the simulation can be obtained by modelling the actuator electrodynamics and the a stable joint torque controller. On a higher level, the perception stack of the robot can be further fully integrated with vision and proprioceptive sensors for multi modal path planning and optimal foot position placement.